\documentclass[journal]{IEEEtran}

\usepackage{amsmath,amssymb}
\usepackage{url}
\usepackage{array}
\usepackage{multirow}
\usepackage{enumerate}
\usepackage{graphicx}
\usepackage{tabularx}
\usepackage{color, colortbl}
\ifCLASSOPTIONcompsoc
  \usepackage[caption=false,font=normalsize,labelfont=sf,textfont=sf]{subfig}
\else
  \usepackage[caption=false,font=footnotesize]{subfig}
\fi

\def\x{{\mathbf x}}
\def\K{\mathbf{K}}

\graphicspath{{./}{Figures/}}
\makeatletter

\newcommand{\Rmnum}[1]{\expandafter\@slowromancap\romannumeral #1@}
\newcolumntype{Y}{>{\centering\arraybackslash}X} % centered X column in tabularx
\definecolor{Gray}{gray}{0.9} % color table cell
\makeatother

\title{Optimizing Kernel Machines using Deep Learning}

\author{Huan~Song,~\IEEEmembership{Member,~IEEE,}
        Jayaraman~J.~Thiagarajan,~\IEEEmembership{Member,~IEEE,} Prasanna~Sattigeri,~\IEEEmembership{Member,~IEEE,}
        and~Andreas~Spanias,~\IEEEmembership{Fellow,~IEEE}% <-this % stops a space
\thanks{Huan Song and Andreas Spanias are with the SenSIP Center, School of Electrical, Computer and Energy Engineering, Arizona State University, Tempe, AZ 85287 USA (e-mail: huan.song@asu.edu and spanias@asu.edu).}% <-this % stops a space
\thanks{Jayaraman J. Thiagarajan is with the Lawrence Livermore National Laboratory, Livermore, CA 94550 USA (email: jjayaram@llnl.gov).}% <-this % stops a space
\thanks{Prasanna Sattigeri is with IBM T. J. Watson Research Center, Yorktown Heights, NY 10598 USA (email: psattig@us.ibm.com).}
}
          
\begin{document}

\maketitle

\begin{abstract}
Building highly non-linear and non-parametric models is central to several state-of-the-art machine learning systems. Kernel methods form an important class of techniques that induce a reproducing kernel Hilbert space (RKHS) for inferring non-linear models through the construction of similarity functions from data. These methods are particularly preferred in cases where the training data sizes are limited and when prior knowledge of the data similarities is available. Despite their usefulness, they are limited by the computational complexity and their inability to support end-to-end learning with a task-specific objective. On the other hand, deep neural networks have become the de facto solution for end-to-end inference in several learning paradigms. In this article, we explore the idea of using deep architectures to perform kernel machine optimization, for both computational efficiency and end-to-end inferencing. To this end, we develop the DKMO (\textit{Deep Kernel Machine Optimization}) framework, that creates an ensemble of dense embeddings using Nystr\"{o}m kernel approximations and utilizes deep learning to generate task-specific representations through the fusion of the embeddings. Intuitively, the filters of the network are trained to fuse information from an ensemble of linear subspaces in the RKHS. Furthermore, we introduce the kernel dropout regularization to enable improved training convergence. Finally, we extend this framework to the multiple kernel case, by coupling a global fusion layer with pre-trained deep kernel machines for each of the constituent kernels. Using case studies with limited training data, and lack of explicit feature sources, we demonstrate the effectiveness of our framework over conventional model inferencing techniques.

%This paper presents a new framework for kernel learning utilizing deep architectures. The connection between kernel representation and deep learning is constructed by multiple dense embeddings obtained through  kernel approximation. Then the deep learning architectures are responsible for learning more expressive representation. Furthermore, we extend the framework for fusing multiple kernels. Experimental results on several real-world datasets show that the proposed frameworks achieve superior classification performance than kernel methods on single kernel learning and multiple kernel fusion tasks. 
\end{abstract}


\IEEEpeerreviewmaketitle

\section{Introduction}
\IEEEPARstart{T}he recent surge in representation learning for complex, high-dimensional data has revolutionized machine learning and data analysis. The success of Deep Neural Networks (DNNs) in a wide variety of computer vision tasks has emphasized the need for highly non-linear and nonparametric models \cite{nair2010rectified,he2016deep}. In particular, by coupling modern deep architectures with large datasets \cite{deng2009imagenet,abu2016youtube}, efficient optimization strategies \cite{ioffe2015batch,srivastava2014dropout} and GPU utilization, one can obtain highly effective predictive models. By using a composition of multiple non-linear transformations, along with novel loss functions, DNNs can approximate a large class of functions for prediction tasks. However, the increasing complexity of the networks requires exhaustive tuning of several hyper-parameters in the discrete space of network architectures, often resulting in sub-optimal solutions or model overfitting. This is particularly more common in applications characterized by limited dataset sizes and complex dependencies in the input space. Despite the advances in regularization techniques and data augmentation strategies \cite{krizhevsky2012imagenet}, in many scenarios, it is challenging to obtain deep architectures that provide significant performance improvements over conventional machine learning solutions. In such cases, a popular alternative solution to building effective, non-linear predictive models is to employ kernel machines. 

\subsection{Kernel Methods and Multiple Kernel Learning}
Kernel methods have a long-standing success in machine learning, primarily due to their well-developed theory, convex formulations, and their flexibility in incorporating prior knowledge of the dependencies in the input space. Denoting the $d-$dimensional input domain as $\mathcal{X} \subset \mathbb{R}^d$, the kernel function $k: \mathcal{X} \times \mathcal{X} \mapsto \mathbb{R}$ induces an implicit mapping into a reproducing kernel Hilbert space (RKHS) $\mathcal{H}_k$, through the construction of a positive definite similarity matrix between samples in the input space. An appealing feature of this approach is that even simple linear models inferred in the RKHS are highly effective compared to their linear counterparts learned directly in the input space. 

Kernel methods are versatile in that specifying a positive-definite kernel will enable the use of this generic optimization framework for any data representation, such as vectors, matrices, sequences or graphs. Consequently, a broad range of kernel construction strategies have been proposed in the literature, e.g. $\chi^2$ kernel \cite{zhang2007local}, string \cite{lodhi2002text}, and graph kernels \cite{vishwanathan2010graph}. Furthermore, the classical Representer Theorem allows the representation of any optimal function in $\mathcal{H}_k$ thereby enabling construction of a dual optimization problem based only on the kernel matrix and not the samples explicitly. This is commonly referred as the \textit{kernel trick} in the machine learning literature. Finally, kernel methods can be augmented with a variety of strategies for controlling the learning capacity and hence reducing model overfitting \cite{cawley2010over}.
%%%% representor theorem equation
%as
%\begin{equation}
%f_{opt}(\x) = \sum_i \alpha_i k(\x,\x_i),
%\label{eqn:rep}
%\end{equation}

Despite these advantages, kernel methods have some crucial limitations when applied in practice: (a) The first limitation is their computational complexity, which grows quadratically with the sample size due to the computation of the kernel (Gram) matrix. A popular solution to address this challenge is to approximate the kernel matrix using the Nystr\"{o}m method \cite{drineas2005nystrom} or the random Fourier features based methods for shift-invariant kernels \cite{rahimi2007random}. While the Nystr\"{o}m method obtains a low-rank approximation of the kernel matrix, the latter explicitly maps the data into an Euclidean inner product space using randomized feature maps; (b) Another crucial limitation of kernel methods is that, unlike the state-of-the-art deep learning systems, the data representation and model learning stages are decoupled and hence cannot admit end-to-end learning. 

%\subsection{Multiple Kernel Learning}
%%%% Deleted this paragraph.
%Similar to many other learning paradigms, it is challenging to choose the right parameters for a specific application, e.g. which kernel to use? and which parameters to use for the chosen kernel? In practice, the suitable parameters are chosen using a grid-search on each of the parameters, coupled with cross-validation. However, this exhaustive search can become quickly intractable, when additional free variables such as the classifier hyperparameters and feature design optimization come into play. 
The active study in multiple kernel learning (MKL) alleviates this limitation to some extent. MKL algorithms \cite{gonen2011multiple} attempt to automatically select and combine multiple base kernels to exploit the complementary nature of the individual feature spaces and thus improve performance. A variety of strategies can be used for combining the kernel matrices, such that the resulting matrix is also positive definite, i.e. a valid kernel. Common examples include non-negative sum \cite{sun2010multiple} or hadamard product of the matrices \cite{li2010nonlinear}. Although MKL provides additional parameters to obtain an optimal RKHS for effective inference, the optimization (dual) is computationally more challenging, particularly with the increase in the number of kernels. More importantly, in practice, this optimization does not produce consistent performance improvements over a simple baseline kernel constructed as the unweighted average of the base kernels \cite{gehler2009feature, cortes2009learning}. Furthermore, extending MKL techniques, designed primarily for binary classification, to multi-class classification problems is not straightforward. In contrast to the conventional one-vs-rest approach, which decomposes the problem into multiple binary classification problems, in MKL it is beneficial to obtain the weighting of base kernels with respect to all classes \cite{zien2007multiclass,cortes2013multi}. 

\subsection{Bridging Deep Learning and Kernel Methods}

In this paper, our goal is to utilize deep architectures to facilitate improved optimization of kernel machines, particularly in scenarios with limited labeled data and prior knowledge of the relationships in data (e.g. biological datasets). Existing efforts on bridging deep learning with kernel methods focus on using kernel compositions to emulate neural network layer stacking or enabling the optimization of deep architectures with data-specific kernels \cite{song2017deep}. Combining the advantages of these two paradigms of predictive learning has led to new architectures and inference strategies. For example, in \cite{mairal2014convolutional,mairal2016end}, the authors utilized kernel learning to define a new type of convolutional networks and demonstrated improved performance in inverse imaging problems (details in Section \ref{Sec:kernellearn}).

%%%% Deleted the paragraph below, previously at the end of Section I-A Kernel Methods in Machine Learning.
%In this paper, we propose to utilize deep neural networks to enable end-to-end optimization of kernel machines and obtain better predictive models in scenarios where kernel methods are typically employed. To this end, we create a dense embedding layer for input samples based on their kernel matrices, and employ end-to-end deep learning with novel regularization strategies to perform predictive model inference. 

Inspired by these efforts, in this paper, we develop a deep learning based solution to kernel machine optimization, for both single and multiple kernel cases. While existing kernel approximation techniques make kernel learning efficient, utilizing deep networks enables end-to-end inference with a task-specific objective. In contrast to approaches such as \cite{mairal2014convolutional,mairal2016end}, which replace the conventional neural network operations, e.g. convolutions, using equivalent computations in the RKHS, we use the similarity kernel to construct dense embeddings for data and build task-specific representations, through fusion of these embeddings. Consequently, our approach is applicable to any kind of data representation. Similar to conventional kernel methods, our approach exploits the native space of the chosen kernel during inference, thereby controlling the capacity of learned models, and thus leading to improved generalization. Finally, in scenarios where multiple kernels are available during training, either corresponding to multiple feature sources or from different kernel parameterizations, we develop a multiple kernel variant of the proposed approach. Interestingly, in scenarios with limited amounts of data and in applications with no access to explicit feature sources, the proposed approach is superior to state-of-the-practice kernel machine optimization techniques and deep feature fusion techniques.
%\subsection{Summary of Contributions} 

The main contributions of this work can be summarized as follows:

\begin{itemize}
\item We develop \textit{Deep Kernel Machine Optimization} (DKMO), which creates dense embeddings for the data through projection onto a subspace in the RKHS and learns task-specific representations using deep learning.

\item To improve the effectiveness of the representations, we propose to create an ensemble of embeddings obtained from Nystr\"{o}m approximation methods, and pose the representation learning task as deep feature fusion.

\item We introduce the kernel dropout regularization to enable robust feature learning with kernels from limited data.

\item We develop M-DKMO, a multiple kernel variant of the proposed algorithm, to effectively perform multiple kernel learning with multiple feature sources or kernel parameterizations.

\item We show that on standardized datasets, where kernel methods have had proven success, the proposed approach outperforms state-of-the-practice kernel methods, with a significantly simpler optimization.

\item Using cell biology datasets, we demonstrate the effectiveness of our approach in cases where we do not have access to features but only encoded relationships.

\item Under the constraint of limited training data, we show that our approach outperforms both the state-of-the-art MKL methods and standard deep neural networks applied to the feature sources directly.

\end{itemize}

\section{Related Work}
\label{Sec:related}
In this section, we briefly review the prior art in optimizing kernel machines and discuss the recent efforts towards bridging kernel methods and deep learning.

\subsection{Kernel Machine Optimization}
\label{Sec:keropt}

%%%% Previously from intro.
%Denoting the $d-$dimensional input domain as $\mathcal{X} \subset \mathbb{R}^d$, the kernel function $k: \mathcal{X} \times \mathcal{X} \mapsto \mathbb{R}$ induces a RKHS $\mathcal{H}_k$ with the corresponding inner product $<.,.>_{\mathcal{H}_k}$ and the norm $\|.\|_{\mathcal{H}_k}$. For a set of data-label pairs $\{\x_i, y_i \}_{i=1}^n$, where $y_i$ corresponds to the label of the sample $\mathbf{x}_i \in \mathbb{R}^d$, the problem of inferring a predictive model can be posed as the following empirical risk minimization task \cite{andrew2000introduction}:
%\begin{equation}
%f_{opt} = \argmin_{f \in \mathcal{H}_k} \frac{1}{n} \sum_i \mathcal{L}(y_i, f(\x_i)) + \lambda \|f\|_{\mathcal{H}_k},
%\label{eq:reglearn}
%\end{equation}where $\mathcal{L}$ denotes a chosen loss function and $\lambda$ is the regularization parameter. For example, in kernel ridge regression $\mathcal{L}$ is chosen to be the $\ell_2$ loss while kernel Support Vector Machine (SVM) uses the hinge loss. A variety of general purpose kernels are used in practice, such as the polynomial and radial basis function (RBF) kernels.
%%%% Above were from previous intro.

The success of kernel Support Vector Machines (SVM) \cite{andrew2000introduction} motivated the kernelization of a broad range of linear machine learning formulations in the Euclidean space. Popular examples are regression \cite{takeda2007kernel}, clustering \cite{Dhillon2004}, unsupervised and supervised dimension reduction algorithms \cite{yan2007graph}, dictionary learning for sparse representations \cite{Thiagarajan2014} and many others. Following the advent of more advanced data representations in machine learning algorithms, such as graphs and points on embedded manifolds, kernel methods provided a flexible framework to perform statistical learning with such data. Examples include the large class of graph kernels \cite{vishwanathan2010graph} and Grassmannian kernels for Riemannian manifolds of linear subspaces \cite{Harandi2004}.

%Broadly speaking, kernel methods circumvent the challenge of constructing explicit maps to the RKHS by solving the dual formulation of risk minimization problems. 
Despite the flexibility of this approach, the need to deal with kernel matrices makes the optimization infeasible in large scale data. There are two class of approaches commonly used by researchers to alleviate this challenge. First, kernel approximation strategies can be used to reduce both computational and memory complexity of kernel methods, e.g. the Nystr\"{o}m method \cite{drineas2005nystrom}. The crucial component in Nystr\"{o}m kernel approximation strategies is to select a subset of the kernel matrix to recover the inherent relationships in the data. A straightforward uniform sampling on columns of kernel matrix has been demonstrated to provide reasonable performance in many cases \cite{kumar2012sampling}. However, in \cite{zhang2010clustered}, the authors proposed an improved variant of Nystr\"{o}m approximation, that employs distance-based clustering to obtain landmark points in order to construct a subspace in the RKHS. Interestingly, the authors proved that the approximation error is bounded by the quantization error of coding each sample using its closest landmark. In \cite{kumar2009ensemble}, Kumar \textit{et. al.} generated an ensemble of approximations by repeating Nystr\"{o}m random sampling multiple times for improving the quality of the approximation. 

Second, in the case of shift-invariant kernels, random Fourier features can be used to design scalable kernel machines \cite{Rahimi2008, Le2013}. Instead of using the implicit feature mapping in the kernel trick, the authors in \cite{Rahimi2008} proposed to utilize randomized features for approximating kernel evaluation. The idea is to explicitly map the data to an Euclidean inner product space using randomized feature maps, such that kernels can be approximated using Euclidean inner products. Using random Fourier features, Huang \textit{et. al.} \cite{Huang2014} showed that shallow kernel machines matched the performance of deep networks in speech recognition, while being computationally efficient.

\vspace{0.1in}
\noindent \textbf{Combining Multiple Kernels:} A straightforward extension to kernel learning is to consider multiple kernels. Here, the objective is to learn a combination of base kernels $k_1,...,k_M$ and perform empirical risk minimization simultaneously. Conical \cite{sun2010multiple} and convex combinations \cite{rakotomamonjy2008simplemkl} are commonly considered and efficient optimizers such as Sequential Minimal Optimization (SMO) \cite{sun2010multiple} and Spectral Projected Gradient (SPG) \cite{jain2012spf} techniques have been developed. In an extensive review of MKL algorithms \cite{gonen2011multiple}, G\"{o}nen \textit{et al}. showed that the formulation in \cite{cortes2009learning} achieved consistently superior performance on several binary classification tasks. MKL algorithms have been applied to a wide-range of machine learning problems. With base kernels constructed from distinct features, MKL can be utilized as a feature fusion mechanism \cite{gehler2009feature,natarajan2012multimodal,bucak2014multiple,liu2014multiple,song2016auto}. When base kernels originate from different feature sources or kernel parameterizations, MKL automates the kernel selection and parameter tuning process \cite{sun2010multiple,cortes2013multi}. Most recent research in MKL focus on improving the multi-class classification performance \cite{cortes2013multi} and effectively handling training convergence and complexity \cite{orabona2012multi}. 
%In \cite{liu2014multiple}, the authors solved the multiple kernel learning problem directly using its primal formulation, with random Fourier features.

These simple fusion schemes have been generalized further to create localized multiple kernel learning (LMKL) \cite{gonen2008localized, kannao2016tv,moeller2016unified} and non-linear MKL algorithms. In \cite{moeller2016unified}, Moeller \textit{et. al.}. have formulated a unified view of LMKL algorithms:
\begin{equation}
\label{Eq:lmkl}
k_\beta(\x_i,\x_j)=\sum_m \beta_m(\x_i,\x_j)k_m(\x_i,\x_j),
\end{equation}
where $\beta_m$ is the gating function for kernel function $k_m$. In contrast to ``global'' MKL formulations where the weight $\beta_m$ is constant across data, the gating function in Equation (\ref{Eq:lmkl}) takes the data sample as an independent variable and is able to characterize the underlying local structure in data. Several LMKL algorithms differ in how $\beta_m$ is constructed. For example, in \cite{gonen2008localized} $\beta_m$ is chosen to be separable into softmax functions. On the other hand, non-linear MKL algorithms are based on the idea that non-linear combination of base kernels could provide richer and more expressive representations compared to linear mixing. For example,  \cite{cortes2009learning} considered polynomial combination of base kernels and \cite{zhuang2011two} utilized a two-layer neural network to construct a RBF kernel composition on top of the linear combination. 

\subsection{Combining Deep Learning with Kernel Methods}
\label{Sec:kernellearn}

While the recent focus of research in kernel learning has been towards scaling kernel optimization and MKL, there is another important direction aimed at improving the representation power of kernel machines. %by incorporating principles from the state-of-the-art representation learning paradigms. 
In particular, inspired by the exceptional power of deep architectures in feature design and end-to-end learning, a recent wave of research efforts attempt to incorporate ideas from deep learning into kernel machine optimization \cite{cho2009kernel,zhuang2011two,wiering2014multi,wilson2016deep,mairal2016end}. One of the earliest approaches in this direction was developed by Cho \textit{et. al} \cite{cho2009kernel}, in which a new arc-cosine kernel was defined. Based on the observation that arc-cosine kernels possess characteristics similar to an infinite single-layer threshold network, the authors proposed to emulate the behavior of DNN by composition of arc-cosine kernels. The kernel composition idea using neural networks was then extended to MKL by \cite{zhuang2011two}. The connection between kernel learning and deep learning can also be drawn through Gaussian processes as demonstrated in \cite{wilson2016deep}, where Wilson \textit{et al}. derived deep kernels through the Gaussian process marginal likelihood. Another class of approaches directly incorporated kernel machines into Deep Neural Network (DNN) architectures. For example, Wiering \textit{et al}. \cite{wiering2014multi} constructed a multi-layer SVM by replacing neurons in multi-layer perceptrons (MLP) with SVM units. More recently, in \cite{mairal2016end}, kernel approximation is carried out using supervised subspace learning in the RKHS, and backpropagation based training similar to convolutional neural network (CNN) is adopted to optimize the parameters. The experimental results on image reconstruction and super-resolution showed that the new type of network achieved competitive and sometimes improved performance as compared to CNN.

In this paper, we provide an alternative viewpoint to kernel machine optimization by considering the kernel approximate mappings as embeddings of the data and employ deep neural networks to infer task-specific representations as a fusion of an ensemble of subspace projections in the RKHS. Crucial advantages of our approach are that extension to the multiple kernel case is straightforward, and it can be highly robust to smaller datasets. 

\section{Deep Kernel Machine Optimization - Single Kernel Case}
\label{Sec:singlelearn}
\begin{figure}[t]
	\centering
	\includegraphics[width=\linewidth]{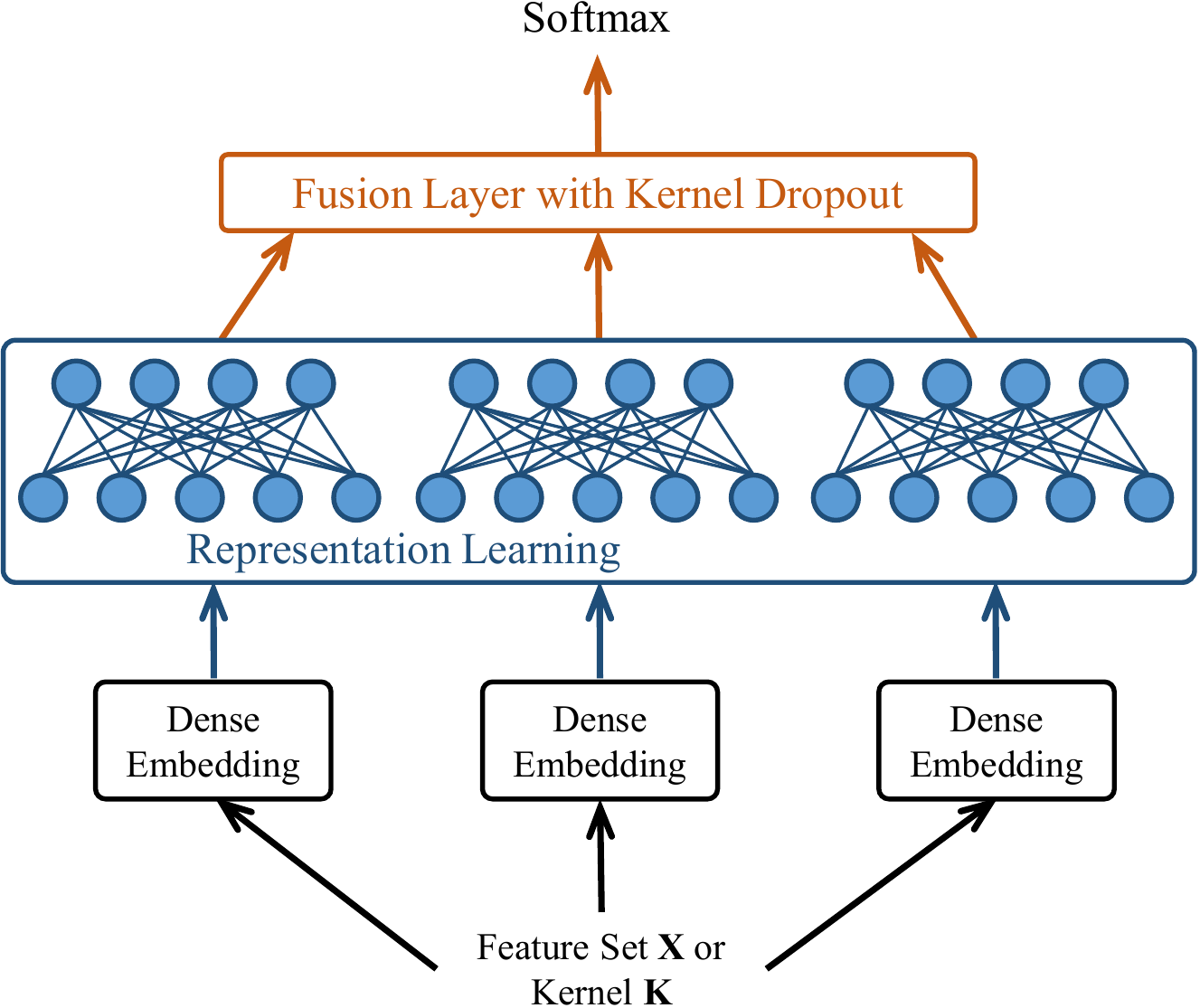}
	\caption{\textit{DKMO} - Proposed approach for optimizing kernel machines using deep neural networks. For a given kernel, we generate multiple dense embeddings using kernel approximation techniques, and fuse them in a fully connected deep neural network. The architecture utilizes fully connected networks with kernel dropout regularization during the fusion stage. Our approach can handle scenarios when both the feature sources and the kernel matrix are available during training or when only the kernel similarities can be accessed.}
	\label{fig:framework1}
\end{figure}  

In this section, we describe the proposed DKMO framework which utilizes the power of deep architectures in end-to-end learning and feature fusion to facilitate kernel learning. Viewed from bottom to top in Figure \ref{fig:framework1}, DKMO first extracts multiple dense embeddings from a precomputed similarity kernel matrix $\mathbf{K}$ and optionally the feature source $\mathcal{X}$ if accessible during training. On top of each embedding, we build a fully connected neural network for representation learning. Given the inferred latent spaces from representation learning, we stack a fusion layer which is responsible for combining the latent features and obtaining a concise representation for inference tasks. Finally, we use a softmax layer at the top to perform classification, or an appropriate dense layer for regression tasks. Note that, similar to random Fourier feature based techniques in kernel methods, we learn a mapping to the Euclidean space, based on the kernel similarity matrix. However, in contrast, the representation learning phase is not decoupled from the actual task, and hence can lead to higher fidelity predictive models. 

\subsection{Dense Embedding Layer}
\label{sec:dense_emb}

From Figure \ref{fig:framework1}, it can be seen that the components of representation learning and fusion of hidden features are generic, i.e., they are separate from the input data or the kernel. Consequently, the dense embedding layer is the key component that bridges kernel representations with the DNN training, thereby enabling an end-to-end training. 

\vspace{0.1in}
\noindent \textbf{Motivation: }Consider the kernel Gram matrix $\mathbf{K}\in\mathbb{R}^{n\times n}$, where $\mathbf{K}_{i,j}=k(\mathbf{x}_i,\mathbf{x}_j)$. The $j$-th column encodes the relevance between sample $\mathbf{x}_j$ to all other samples $\x_i$ in the training set, and hence this can be viewed as an embedding for $\mathbf{x}_j$. As a result, these naive embeddings can potentially be used in the input layer of the network. However, $\mathbf{k}_j$ has large values at location corresponding to training samples belonging to the same class as $\x_j$ and small values close to zero at others. The sparsity and high dimensionality of these embeddings make them unsuitable for inference tasks. 

A natural approach to alleviate this challenge is to adopt kernel matrix factorization strategies, which transform the original embedding into a more tractable, low-dimensional representation. This procedure can be viewed as kernel approximation with truncated SVD or Nystr\"{o}m methods \cite{drineas2005nystrom}. Furthermore, this is conceptually similar to the process of obtaining dense word embeddings in natural language processing. For example, Levy \textit{et.al} \cite{levy2014neural} have showed that the popular skip-gram with negative sampling (SGNS) model in language modeling is implicitly factorizing the Pointwise Mutual Information matrix, whose entries measure the association between pairs of words. Interestingly, they demonstrated that alternate word embeddings obtained using the truncated SVD method are more effective than SGNS on some word modeling tasks \cite{levy2014neural}.

In existing deep kernel learning approaches such as the convolutional kernel networks \cite{mairal2016end}, the key idea is to construct multiple reproducing kernel Hilbert spaces at different convolutional layers of the network, with a sequence of pooling operations between the layers to facilitate kernel design for different sub-region sizes. However, this approach cannot generalize to scenarios where the kernels are not constructed from images, for example, in the case of biological sequences. Consequently, we propose to obtain multiple approximate mappings (dense embeddings) from the feature set or the kernel matrix using Nystr\"{o}m methods, and then utilize the DNN as both representation learning and feature fusion mechanisms to obtain a task-specific representation for data in the Euclidean space. All components in this framework are general and are not constrained by the application or kind of data used for training.

\vspace{0.1in}
\noindent \textbf{Dense Embeddings using Nystr\"{o}m Approximation:} In order to be flexible with different problem settings, we consider two different pipelines for constructing the dense embeddings based on Nystr\"{o}m approximation: \Rmnum 1) In many applications, e.g. biological sequences or social networks, it is often easier to quantify sample-to-sample distance or similarity than deriving effective features or measurements for inference tasks. Furthermore, for many existing datasets, large-scale pair-wise distances are already pre-computed and can be easily converted into kernel matrices. In such scenarios, we use the conventional Nystr\"{o}m method to calculate the dense embeddings. \Rmnum 2) When the input data is constructed from pre-defined feature sources, we employ the clustered Nystr\"{o}m method \cite{zhang2010clustered}, which identifies a subspace in the RKHS using distance-based clustering, and explicitly project the feature mappings onto subspaces in the RKHS. In this case, the dense embeddings are obtained without constructing the complete kernel matrix for the dataset. Next, we discuss these two strategies in detail.

%\begin{equation}
%\mathbf{U_K}\simeq\sqrt{\frac{r}{n}}\mathbf{E_Z} \mathbf{U_Z}\mathbf{\Lambda}_\mathbf{Z}^{-1},~~\mathbf{\Lambda_K}=\frac{n}{r}\mathbf{\Lambda_Z}
%\end{equation}
%we approximate $\mathbf{U_K}$, $\mathbf{\Lambda_K}$ by the eigenvectors $\mathbf{U_Z}$ and the associated eigenvalues $\mathbf{\Lambda_Z}$ of $\mathbf{W}$. Formally,
%$\mathbf{L}\simeq \mathbf{U_K}\mathbf{\Lambda}_\mathbf{Z}^{-1/2}=\mathbf{E_Z}(\mathbf{U_Z}\mathbf{\Lambda}_\mathbf{Z}^{-1/2}).$

\subsubsection{Conventional Nystr\"{o}m approximation on kernels} 
In applications where the feature sources are not directly accessible, we construct dense embeddings from the kernel matrix directly. Based on the Nystr\"{o}m method \cite{kumar2012sampling, kumar2009ensemble}, a subset of $s$ columns selected from $\mathbf{K}$ can be used to find an approximate kernel map $\mathbf{L} \in\mathbb{R}^{n\times r}$, such that $\mathbf{K}\simeq \mathbf{L} \mathbf{L}^T$ where $s\ll n$ and $r\leq s$. To better facilitate the subsequent DNN representation learning, we extract multiple approximate mappings through different random samplings of the kernel matrix. More specifically, from $\mathbf{K}$, we randomly select $s \times P$ columns without replacement, and then divide it into $P$ sets containing $s$ columns each. Consider a single set $\mathbf{E}\in \mathbb{R}^{n\times s}$ containing the selected columns and denote $\mathbf{W}\in \mathbb{R}^{s\times s}$ as the intersection of the selected columns and corresponding rows on $\mathbf{K}$. The rank-$r$ approximation $\tilde{\mathbf{K}}_r$ of $\mathbf{K}$ is computed as 
\begin{equation}
\tilde{\mathbf{K}}_r=\mathbf{E}\tilde{\mathbf{W}}_r \mathbf{E}^T
\end{equation}
where $\tilde{\mathbf{W}}_r$ is the optimal rank-$r$ approximation of $\mathbf{W}$ obtained using truncated SVD. As it can be observed, the time complexity of the approximation reduces to $O(s^3)$, which corresponds to performing SVD on $\mathbf{W}$. This can be further reduced by randomized SVD algorithms as shown in \cite{li2015large}. The approximate mapping function $\mathbf{L}$ can then be obtained by
\begin{equation}
\label{eq:appr1}
\mathbf{L}=\mathbf{E}(\mathbf{U}_{\tilde{\mathbf{W}}_r}\mathbf{\Lambda}_{\tilde{\mathbf{W}}_r}^{-1/2})
\end{equation}
where $\mathbf{U}_{\tilde{\mathbf{W}}_r}$ and $\Lambda_{\tilde{\mathbf{W}}_r}$ are top $r$ eigenvalues and eigenvectors of $\mathbf{W}$.

With different sampling sets spanning distinct subspaces, the projections will result in completely different representations in the RKHS. Since the performance of our end-to-end learning approach is heavily influenced by the construction of subspaces in the RKHS, we propose to infer an ensemble of multiple subspace approximations for a given kernel. This is conceptually similar to \cite{kumar2009ensemble}, in which an ensemble of multiple Nystr\"{o}m approximations are inferred to construct an approximation of the kernel. However, our approach works directly with the approximate mappings $\mathbf{L}$ instead of approximated kernels $\tilde{\mathbf{K}}_r$ and the mappings are further coupled with the DNN optimization. The differences in the representations of the projected features will be exploited in the deep learning fusion architecture to model the characteristics in different regions of the input space. To this end, we repeat the calculation based on Equation (\ref{eq:appr1}) for all $P$ selected sets and obtain the dense embeddings $\mathbf{L}_1,\dots,\mathbf{L}_P$. 

\subsubsection{Clustered Nystr\"{o}m approximation on feature sets} 
When the feature sources are accessible, we propose to employ clustered Nystr\"{o}m approximation to obtain the dense embeddings directly from features without construction of the actual kernel. Following the approach in \cite{zhang2010clustered}, $k$-means cluster centroids can be utilized as the set of the \textit{landmark points} from $\mathbf{X}$. Denoting the matrix of landmark points by $\mathbf{Z}=[\mathbf{z}_1,\dots,\mathbf{z}_r]$ and the subspace they span by $\mathcal{F}=\text{span}(\varphi(\mathbf{z}_1),\dots,\varphi(\mathbf{z}_r))$,  the projection of the samples $\varphi(\x_1),\dots,\varphi(\x_n)$ in $\mathcal{H}_k$ onto its subspace $\mathcal{F}$ is equivalent to the following Nystr\"{o}m approximation (we refer to \cite{mairal2016end} for the detailed derivation): 

\begin{equation}
\label{eq:appr2}
\mathbf{L_Z}=\mathbf{E_Z}\mathbf{W}_\mathbf{Z}^{-1/2}.
\end{equation}where $(\mathbf{E_Z})_{i,j}=k(\mathbf{x}_i,\mathbf{z}_j)$ and $(\mathbf{W_Z})_{i,j}=k(\mathbf{z}_i,\mathbf{z}_j)$. As it can be observed in the above expression, only kernel matrices $\mathbf{W_Z}\in\mathbb{R}^{r\times r}$ and $\mathbf{E_Z}\in\mathbb{R}^{n\times r}$ need to be constructed, which are computationally efficient since $r\ll n$. Note that, comparing Equations (\ref{eq:appr2}) and (\ref{eq:appr1}), $\mathbf{L_Z}$ is directly related to $\mathbf{L}$ by a linear transformation when $r=s$, since
\begin{equation}
\mathbf{W}_\mathbf{Z}^{-1/2}=\mathbf{U_Z}\mathbf{\Lambda}_\mathbf{Z}^{-1/2}\mathbf{U}_\mathbf{Z}^T,
\end{equation}where $\mathbf{U_Z}$ and $\mathbf{\Lambda_Z}$ are eigenvectors and the associated eigenvalues of $\mathbf{W_Z}$ respectively.

Similar to the previous case, we obtain an ensemble of subspace approximations by repeating the landmark selection process with different clustering techniques: the $k$-means, $k$-medians, $k$-medoids, agglomerative clustering \cite{kaufman2009finding} and spectral clustering based on $k$ nearest neighbors \cite{von2007tutorial}. Note that, additional clustering algorithms or a single clustering algorithm with different parameterizations can be utilized as well. For algorithms which only perform partitioning and do not provide cluster centroids (e.g.\ spectral clustering), we calculate the centroid of a cluster as the mean of the features in that cluster. In summary, based on the $P$ different landmark matrices $\mathbf{Z}_1,\dots,\mathbf{Z}_P$, we obtain $P$ different embeddings $\mathbf{L}_1,\dots,\mathbf{L}_P$ for the feature set using Equation (\ref{eq:appr2}).

\begin{figure}[t]
  \centering
  \centerline{\includegraphics[width=\linewidth]{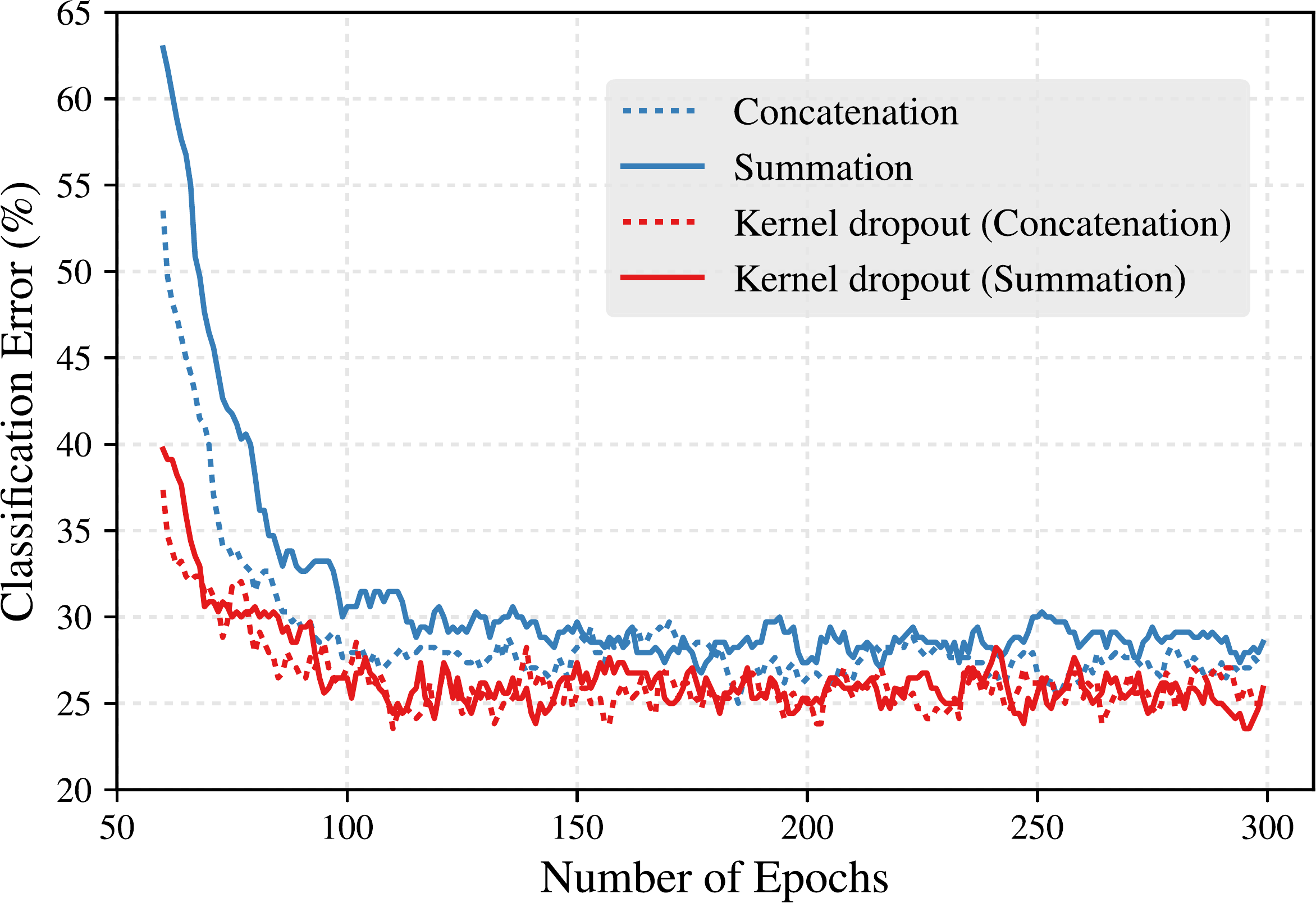}}
\caption{Effects of kernel dropout on the DKMO training process: We compare the convergence characteristics obtained with the inclusion of the kernel dropout regularization in the fusion layer in comparison to the non-regularized version. Note, we show the results obtained with two different merging strategies - concatenation and summation. We observe that the kernel dropout regularization leads to improved convergence and lower classification error for both the merging styles.}
\label{fig:dropout}
\end{figure} 

\subsection{Representation Learning}
Given the kernel-specific dense embeddings, we perform representation learning for each embedding using a multi-layer fully connected network to facilitate the design of a task-specific latent space. Note that, though strategies for sharing weights across the different dense embeddings can be employed, in our implementation we make the networks independent. Following the common practice in deep learning systems, at each hidden layer, dropout regularization \cite{srivastava2014dropout} is used to prevent overfitting and batch normalization \cite{ioffe2015batch} is adopted to accelerate training. 

\subsection{Fusion Layer with Kernel Dropout}
\label{Sec:kerneldropout}
The fusion layer receives the latent representations for each of the RKHS subspace mappings and can admit a variety of fusion strategies to obtain the final representation for prediction tasks. Common merging strategies include concatenation, summation, averaging, multiplication etc. The back propagation algorithm can then be used to optimize for both the parameters of the representation learning and those of the fusion layer jointly to improve the classification accuracy. Given the large number of parameters and the richness of different kernel representations, the training process can lead to overfitting. In order to alleviate this, we propose to impose a \textit{kernel dropout} regularization in addition to the activation dropout in the representation learning phase. 

In the typical dropout regularization \cite{srivastava2014dropout} for training large neural networks, neurons are randomly chosen to be removed from the network along with their incoming and outgoing connections. The process can be viewed as sampling from a large set of possible network architectures with shared weights. In our context, given the ensemble of dense embeddings $\mathbf{L}_1,\dots,\mathbf{L}_P$, an effective regularization mechanism is needed to prevent the network training from overfitting to certain subspaces in the RKHS. More specifically, we propose to regularize the fusion layer by dropping the entire representations learned from some randomly chosen dense embeddings. Denoting the hidden layer representations before the fusion as $\mathcal{H}=\{\mathbf{h}_p\}_{p=1}^P$ and a vector $\mathbf{t}$ associated with $P$ independent Bernoulli trials, the representation $\mathbf{h}_p$ is dropped from the fusion layer if $t_p$ is 0. The feed-forward operation can be expressed as:
\begin{gather*}
t_p\sim \text{Bernoulli}(P) \\
\tilde{\mathcal{H}}=\{\mathbf{h} \mid \mathbf{h} \in \mathcal{H} \text{ and } t_p>0\} \\
\tilde{\mathbf{h}}=(\mathbf{h}_i),\mathbf{h}_i\in \tilde{\mathcal{H}} \\
\tilde{y}_i=f(\mathbf{w}_i\tilde{\mathbf{h}}+b_i),
\end{gather*}where $\mathbf{w}_i$ are the weights for hidden unit $i$, $(\cdot)$ denotes vector concatenation and $f$ is the softmax activation function. In Figure \ref{fig:dropout}, we illustrate the effects of kernel dropout on the convergence speed and classification performance of the network. The results shown are obtained using one of the kernels used in protein subcellular localization (details in Section \ref{sec:exp:only_kernels}). We observe that, for both the merging strategies (concatenation and summation), using the proposed regularization leads to improved convergence and produces lower classification error, thereby evidencing improved generalization of kernel machines trained using the proposed approach.

\section{M-DKMO: Extension to Multiple Kernel Learning}
\label{Sec:multilearn}
\begin{figure}[t]
	\centering
	\includegraphics[width=\linewidth]{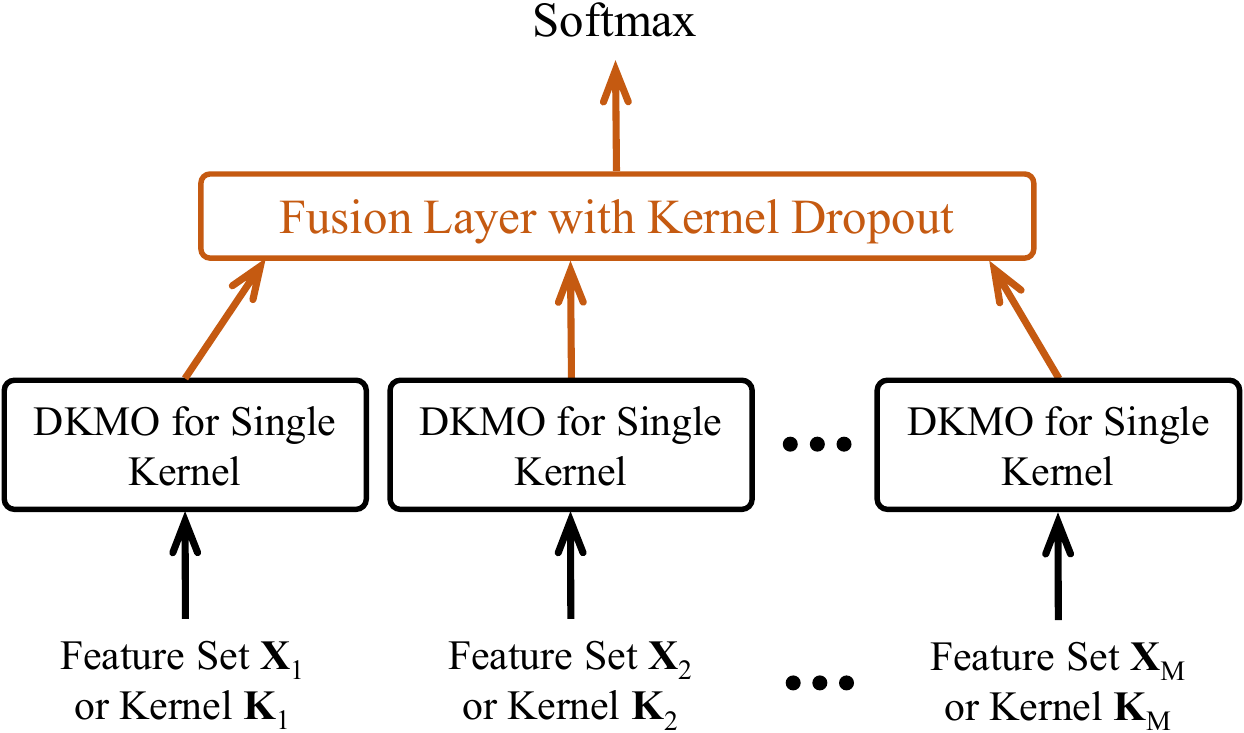}
	\caption{\textit{M-DKMO} - Extending the proposed deep kernel optimization approach to the case of multiple kernels. Each of the kernels are first independently trained with the \textit{DKMO} algorithm in Section \ref{Sec:singlelearn} and then combined using a global fusion layer. The parameters of the global fusion layer and the individual \textit{DKMO} networks are fine-tuned in an end-to-end learning fashion.}
	\label{fig:framework2}
\end{figure}  

As described in Section \ref{Sec:keropt}, extending kernel learning techniques to the case of multiple kernels is crucial to enabling automated kernel selection and fusion of multiple feature sources. The latter is particularly common in complex recognition tasks where the different feature sources characterize distinct aspects of data and contain complementary information. Unlike the traditional kernel construction procedures, the problem of multiple kernel learning is optimized with a task-specific objective, for example hinge loss in classification. In this section, we describe the multiple kernel variant of the deep kernel machine optimization (\textit{M-DKMO}) presented in the previous section.

In order to optimize kernel machines with multiple kernels $\{\mathbf{K}\}_{m=1}^M$ (optionally feature sets $\{\mathbf{X}\}_{m=1}^M$), we begin by employing the DKMO approach to each of the kernels independently. As we will demonstrate with the experimental results, the representations for the individual kernels obtained using the proposed approach produce superior class separation compared to conventional kernel machine optimization (e.g. Kernel SVM). Consequently, the hidden representations from the learned networks can be used to subsequently obtain more effective features by exploiting the correlations across multiple kernels. Figure \ref{fig:framework2} illustrates the \textit{M-DKMO} algorithm for multiple kernel learning. As shown in the figure, an end-to-end learning network is constructed based on a set of pre-trained DKMO models corresponding to the different kernels and a global fusion layer that combines the hidden features from those networks. Similar to the DKMO architecture in Figure \ref{fig:framework1}, the global fusion layer can admit any merging strategy and can optionally include additional fully connected layers before the softmax layer. 
 
Note that, after pre-training the DKMO network for each of the kernels with a softmax layer, we ignore the final softmax layer and use the optimized network parameters to initialize the M-DKMO network in Figure \ref{fig:framework2}. Furthermore, we adopt the kernel dropout strategy described in Section \ref{Sec:kerneldropout} in the global fusion layer before applying the merge strategy. This regularization process guards against overfitting of the predictive model to any specific kernel and provides much improved generalization. From our empirical studies, we observed that both our initialization and regularization strategies enable consistently fast convergence. 

\section{Experimental Results}
\label{Sec:experiments}
\begin{figure}[t!]
	\centering
	\subfloat[Images from different classes in the flowers102 dataset]{\includegraphics[width=0.65\linewidth]{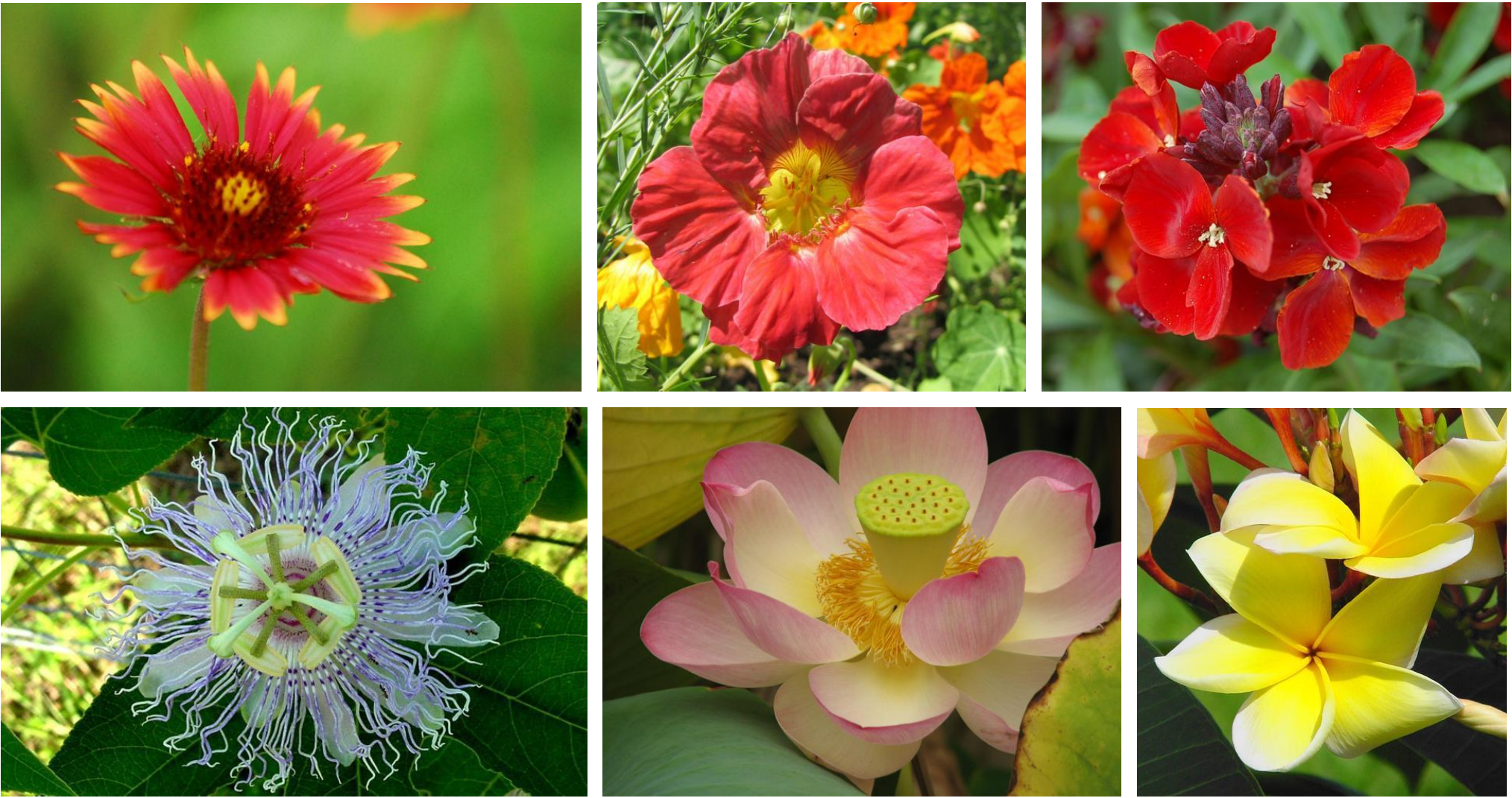}} \\
	\subfloat[Sequences belonging to $3$ different classes in the non-plant dataset for protein subcellular localization]{\includegraphics[width=\linewidth]{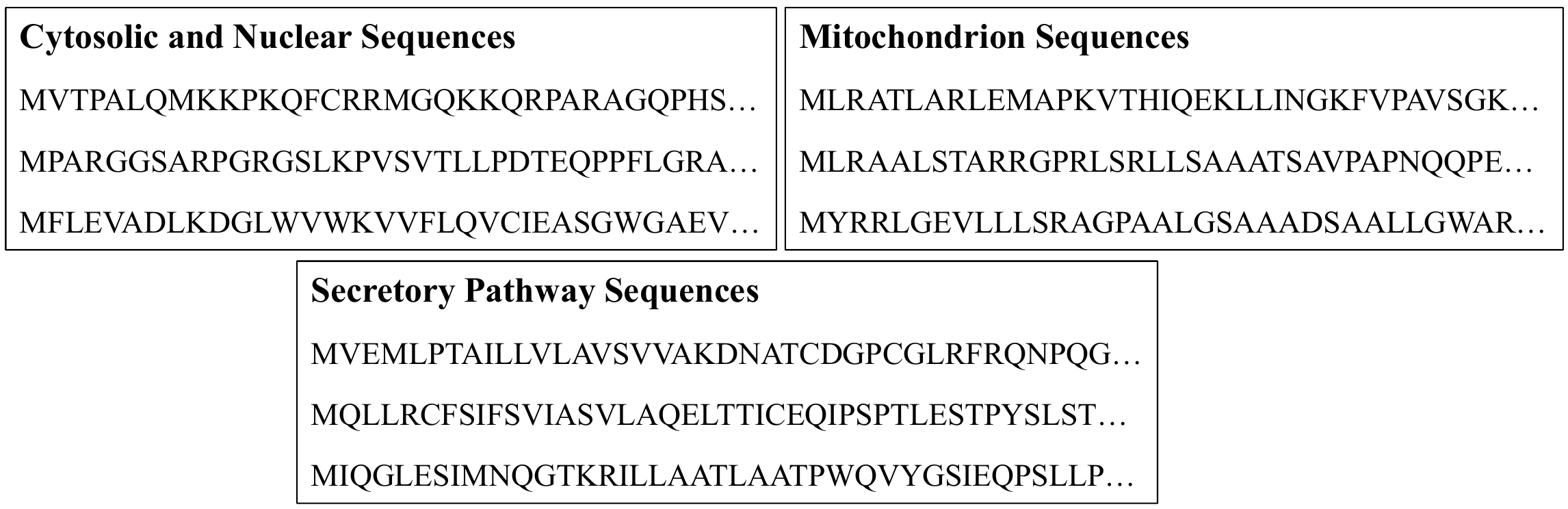}} \\
	\subfloat[Accelerometer measurements characterizing different activities from the USC-HAD dataset] {\includegraphics[width = \linewidth]{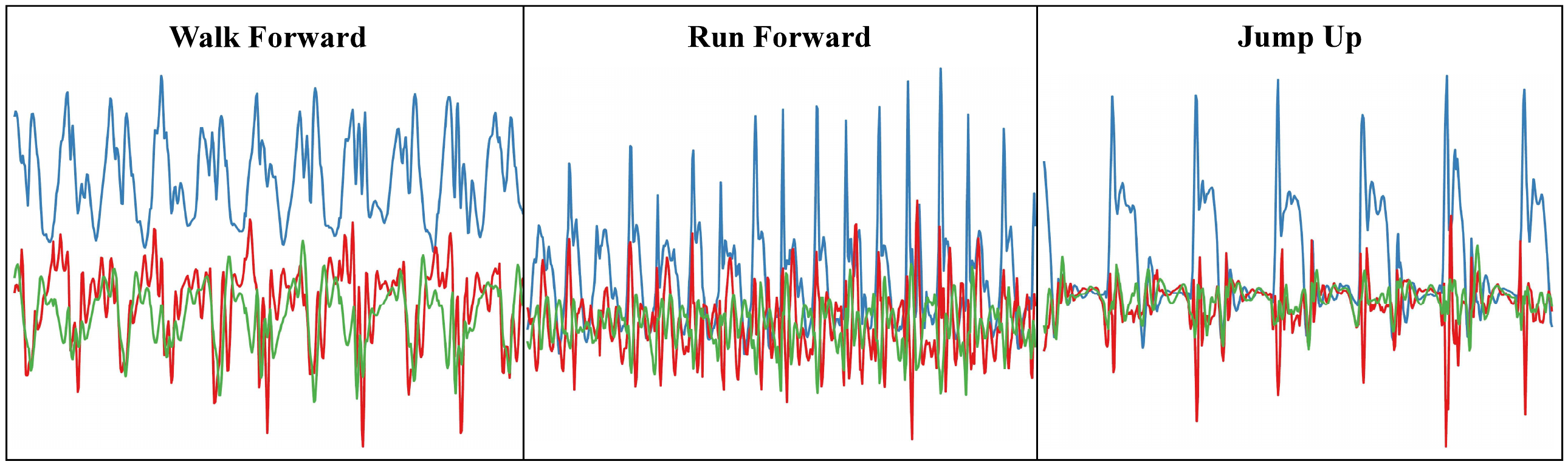}}
	\caption{Example samples from the datasets used in our experiments. The feature sources and kernels are designed based on state-of-the-art practices. The varied nature of the data representations are readily handled by the proposed approach and kernel machines are trained for single and multiple kernel cases.}
	\label{fig:data}
\end{figure}

In this section, we demonstrate the features and performance of the proposed framework using $3$-fold experiments on real-world datasets:

\begin{enumerate}[1)]
\item In Section \ref{sec:exp:with_svm}, we compare with single kernel optimization (kernel SVM) and MKL to demonstrate that the proposed methods are advantageous to the existing algorithms based on kernel methods. To this end, we utilize the standard flowers image classification datasets with pre-computed features. A sample set of images from this dataset are shown in Figure \ref{fig:data}(a).

\item In Section \ref{sec:exp:only_kernels}, we emphasize the effectiveness of proposed architectures when only pair-wise similarities are available from raw data. In subcellular localization, a typical problem in bioinformatics, the data is in the form of protein sequences (as shown in Figure \ref{fig:data}(b)) and as a result, DNN cannot be directly applied for representation learning. In this experiment, we compare with decomposition based feature extraction (\textit{Decomp}) and existing MKL techniques.

\item In Section \ref{sec:exp:limited}, we focus on the performance of the proposed architecture when limited training data is available. As a representative application, sensor based activity recognition requires often laborious data acquisition from human subjects. The difficulty in obtaining large amounts of clean and labeled data can be further complicated by sensor failure, human error and incomplete coverage of the demographic diversity \cite{kwapisz2011activity, zhang2012usc, song2016consensus}. Therefore, it is significant to have a model which has strong extrapolation ability given even very limited training data. When features are accessible, an alternative general-purpose algorithm is fully connected neural networks (\textit{FCN}) coupled with feature fusion. In this section, we compare the performance of our approach with both \textit{FCN} and state-of-the-art kernel learning algorithms. A demonstrative set of time-varying measurements are presented in Figure \ref{fig:data}(c).

\end{enumerate}

\begin{figure*}[!ht]
	\centering
	\subfloat[Flowers17]{\includegraphics[width=0.36\linewidth]{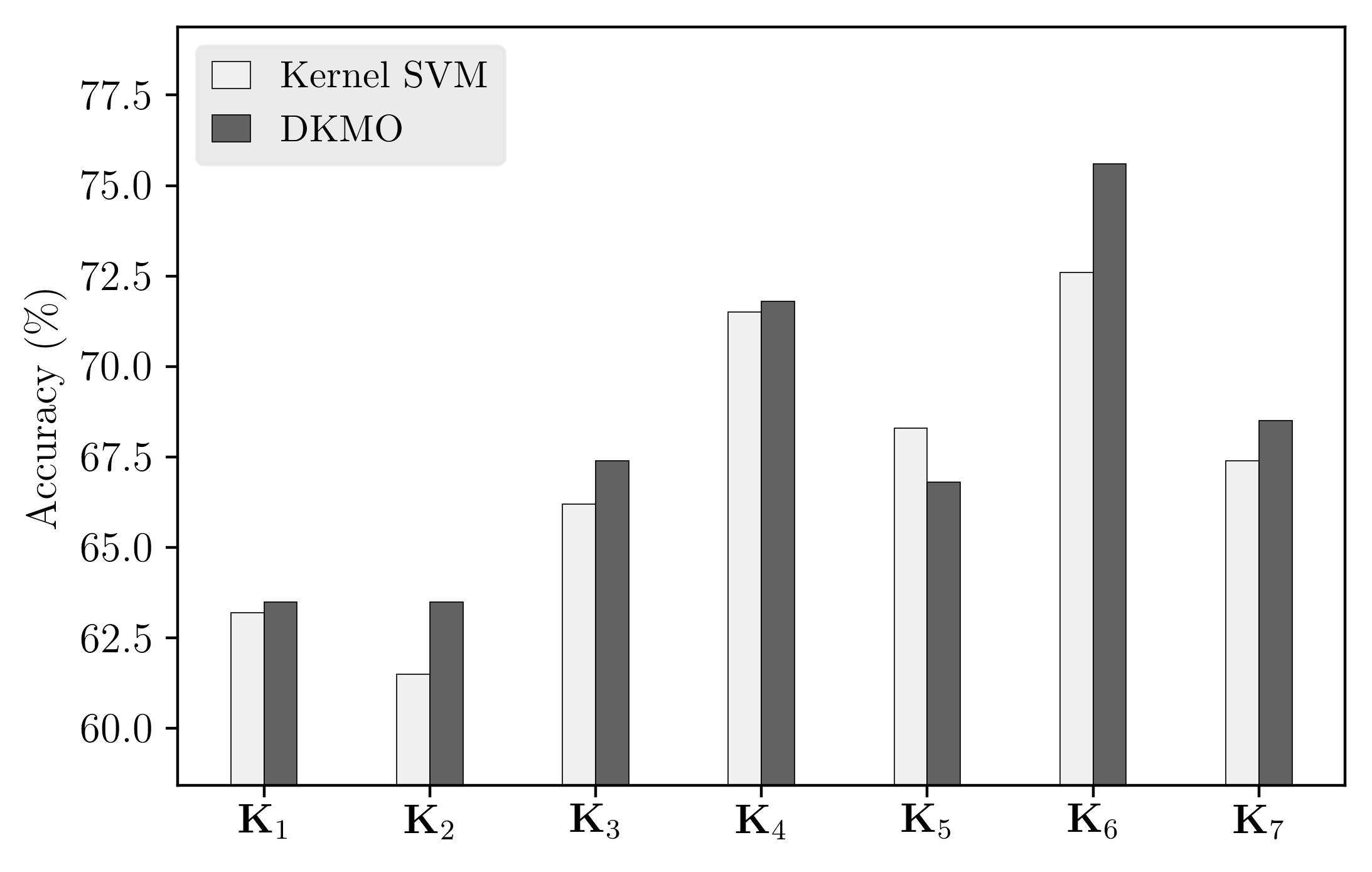}}
	\label{fig:flowers1}
	\subfloat[Flowers102 - 20]{\includegraphics[width=0.25\linewidth]{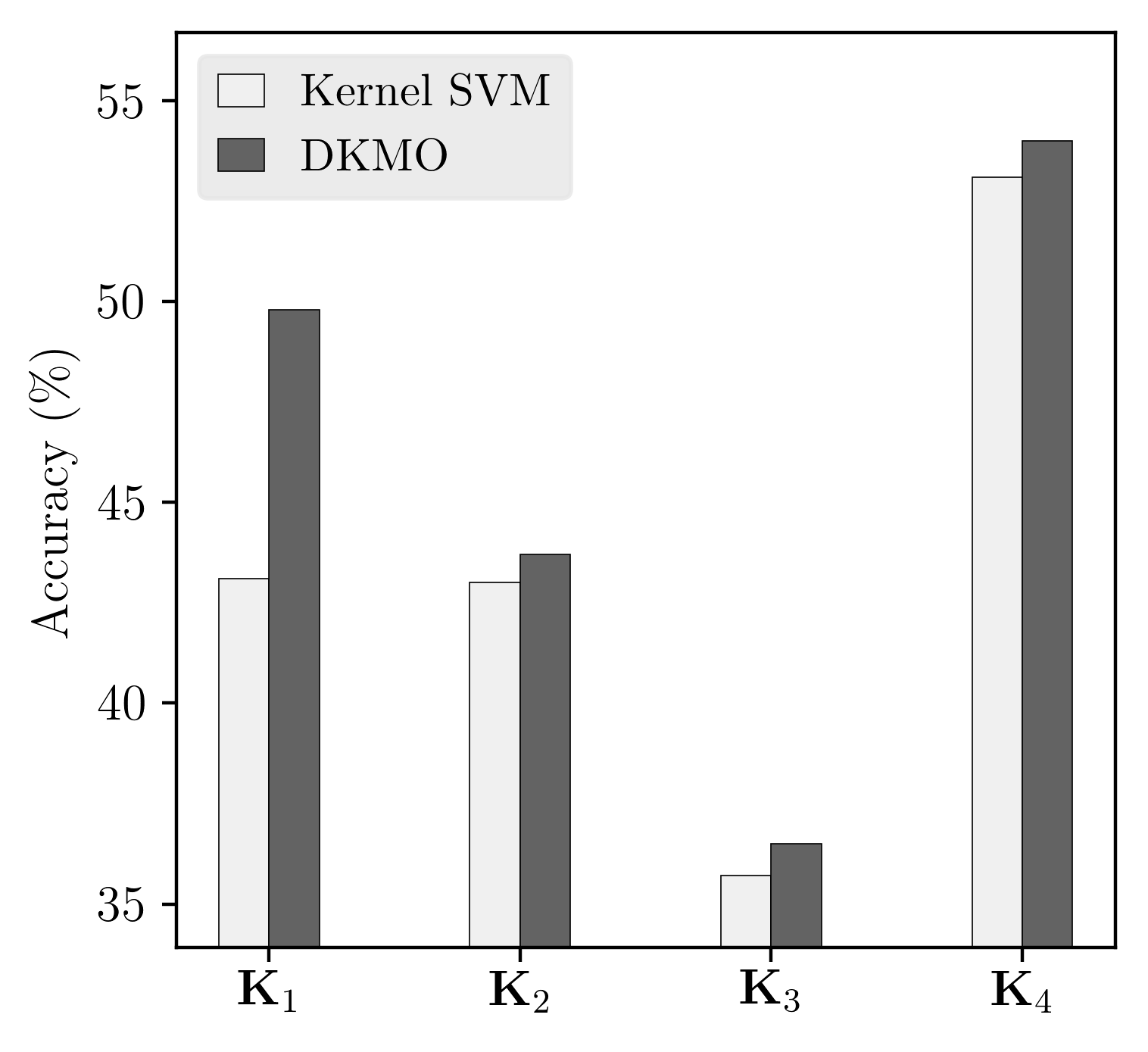}}
	\label{fig:flowers2}
	\subfloat[Flowers102 - 30]{\includegraphics[width=0.25\linewidth]{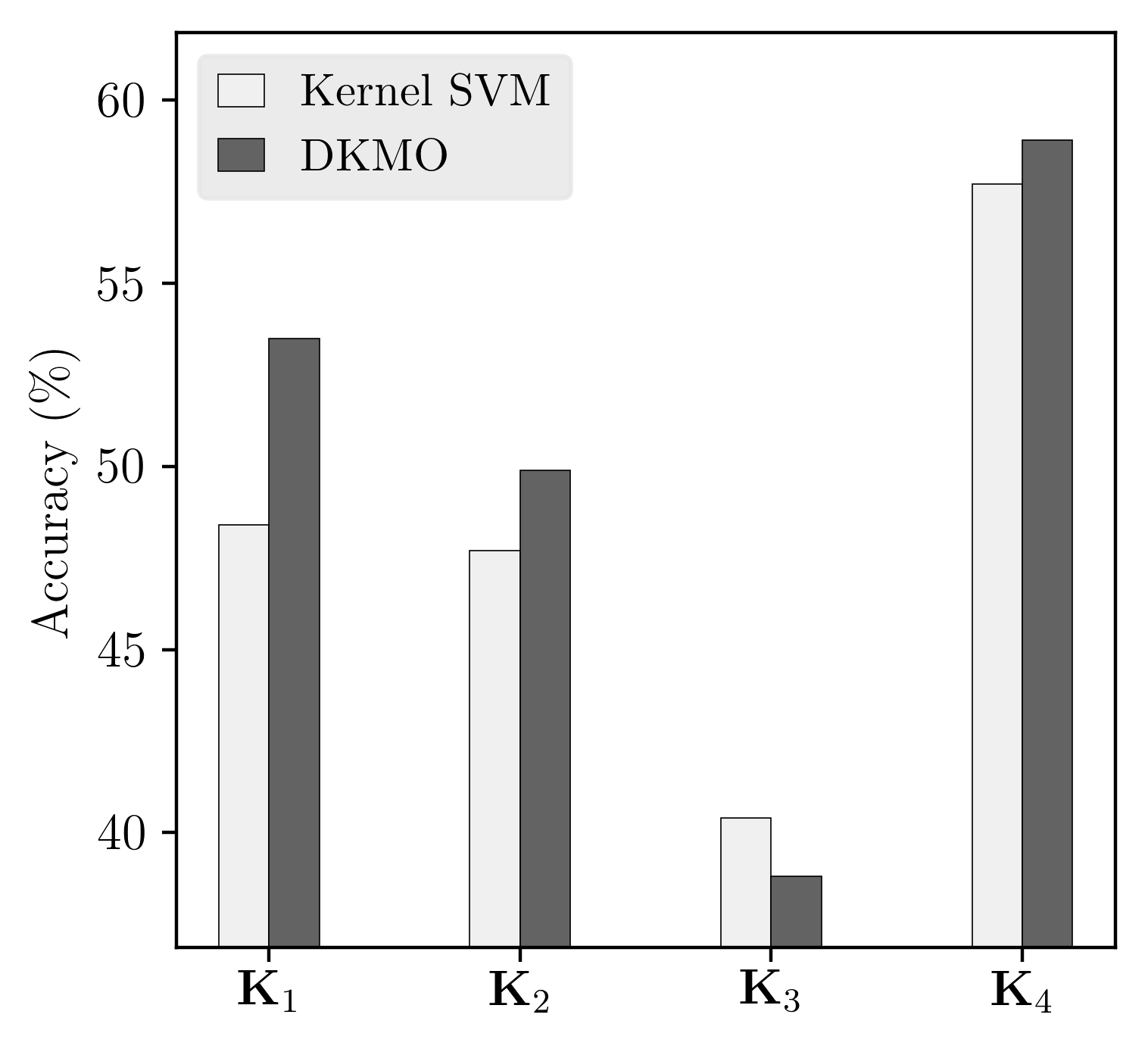}}
	\label{fig:flowers3}
	\caption{Single Kernel Performance on Flowers Datasets}
	\label{fig:res:flowers}
\end{figure*}

As can be seen, the underlying data representations considered in our experiments are vastly different, i.e., images, biological sequences and time-series respectively. The flexibility of the proposed approach enables its use in all these cases without additional pre-processing or architecture fine-tuning. Besides, depending on the application we might have access to the different feature sources or to only the kernel similarities. As described in Section \ref{sec:dense_emb}, the proposed DKMO algorithm can handle both these scenarios by constructing the dense embeddings suitably. 

We summarize all methods used in our comparative studies and the details of the parameters used in our experiments below:

\textit{Kernel SVM}. A single kernel SVM is applied on each of the kernels. Following \cite{chapelle2005semi}, the optimal $C$ parameters for kernel SVM were obtained based on a grid search on $[10^{-1},10^0,10^1,10^2]\times C^{*}$ through cross-validation on the training set, where the default value $C^{*}$ was calculated as $C^{*}=1/(\frac{1}{n}\sum_i \K_{i,i}-\frac{1}{n^2}\sum_{ij} \K_{i,j})$, which is the inverse of the empirical variance of data in the input space. 

\textit{Uniform}. Simple averaging of base kernels has been shown to be a strong baseline in comparison to MKL \cite{gehler2009feature, cortes2009learning}. We then apply \textit{kernel SVM} on the averaged kernel.

\textit{UFO-MKL}. We compare with this state-of-the-art multiple kernel learning algorithm \cite{orabona2011ultra}. The optimal $C$ parameters were cross-validated on the grid $[10^{-1},10^0,10^1,10^2,10^3]$.

\textit{Decomp}. When only kernel similarities are directly accessible (Section \ref{sec:exp:only_kernels}), we compute decomposition based features using truncated SVD. A linear SVM is then learned on the features with similar parameter selection procedure as in \textit{kernel SVM}.

\textit{Concat}. In order to extend \textit{Decomp} to the multiple kernel case, we concatenate all \textit{Decomp} features before learning a classifier.

\textit{FCN}. We construct a fully connected network for each feature set (using \textit{Decomp} feature if only kernels are available) consisting of $4$ hidden layers with sizes $256-512-256-128$ respectively. For the multiple kernel case, a concatenation layer merges all \textit{FCN} built on each set. In the training process, batch normalization and dropout with fixed rate of $0.5$ are used after every hidden layer. The optimization was carried out using the Adam optimizer, with the learning rate set at $0.001$. 

\textit{DKMO} and \textit{M-DKMO}. For all the datasets, we first applied the DKMO approach to each of the kernels (as in Figure \ref{fig:framework1}) with the same network size as in \textit{FCN}. Based on the discussion in Section \ref{sec:dense_emb}, for datasets that allow access to explicit feature sources, we extracted $5$ dense embeddings corresponding to the $5$ landmark point sets obtained using different clustering algorithms. On the other hand, for datasets with only kernel similarity matrices between the samples, we constructed $6$ different dense embeddings with varying subset sizes and approximation ranks. We performed kernel dropout regularization with summation merging for the fusion layer in the DKMO architecture. The kernel dropout rate was fixed at $0.5$. For multiple kernel fusion using the M-DKMO approach, we normalize each kernel as $\bar{\mathbf{K}}_{i,j}=\mathbf{K}_{i,j}/\sqrt{\K_{i,i}\K_{j,j}}$, so that $\bar{\K}_{i,i}=1$. Similar to the DKMO case, we set the kernel dropout rate at $0.5$ and used summation based merging at the global fusion layer in M-DKMO. Other network learning parameters were same as the ones in the \textit{FCN} method. All network architectures were implemented using the Keras library \cite{chollet2015keras} with the TensorFlow backend and trained on a single GTX 1070 GPU. 

\subsection{Image Classification - Comparisons with Kernel Optimization and Multiple Kernel Learning}
\label{sec:exp:with_svm}

% Original table for flowers datasets
%\begin{table*}[!t]
%	\caption{Classification Performance on Flowers Datasets}
%	\label{table:img}
%	\setlength\tabcolsep{2.5pt}
%	\centering
%	\begin{tabularx}{0.8\textwidth}{ |c Y Y Y Y Y Y Y | c c c |}
%		\hline
%		\multicolumn{8}{|c|}{\textbf{Single Kernel Learning}} & \multicolumn{3}{c|}{\textbf{Multiple Kernel Learning}} \\
%		Method & Kernel 1 & Kernel 2 & Kernel 3 & Kernel 4 & Kernel 5 & Kernel 6 & Kernel 7 & Uniform & UFO-MKL & M-DKMO \\
%		\hline
%		\hline
%		\rowcolor{Gray}
%		\multicolumn{11}{|c|}{\textsc{Flowers17}, $n=1360$} \\
%		\hline
%		Kernel SVM & 63.2 & 61.5 & 66.2 & 71.5 & 68.3 & 72.6 & 67.4 & \multirow{2}{*}{85.3} & \multirow{2}{*}{87.1} & \multirow{2}{*}{\textbf{90.0}}\\
%		DKMO & \textbf{63.5} & \textbf{63.5} & 62.4 & \textbf{71.8} & 64.1 & \textbf{75.6} & \textbf{67.4} & & & \\
%		\hline
%		\rowcolor{Gray}
%		\multicolumn{11}{|c|}{\textsc{Flowers102 - 20}, $n=8189$} \\
%		\hline
%		Kernel SVM & 43.1 & 43.0 & 35.7 & 53.1 &  - &  - & - &  \multirow{2}{*}{69.9} & \multirow{2}{*}{75.7} & \multirow{2}{*}{\textbf{76.5}} \\
%		DKMO & \textbf{49.8} & \textbf{43.7} & \textbf{36.5} & \textbf{54.0} &  - & -  & - & & & \\
%		\hline
%		\rowcolor{Gray}
%		\multicolumn{11}{|c|}{\textsc{Flowers102 - 30}, $n=8189$} \\
%		\hline
%		Kernel SVM & 48.4 & 47.7 & 40.4 & 57.7 & - & - & - & \multirow{2}{*}{73.0} & \multirow{2}{*}{80.4} & \multirow{2}{*}{\textbf{80.7}} \\
%		DKMO & \textbf{53.5} & \textbf{49.9} & 38.8 & \textbf{58.9} & - & - & - & & & \\
%		\hline
%	\end{tabularx}
%\end{table*}

In this section, we consider the performance of the proposed approach in image classification tasks, using datasets which have had proven success with kernel methods. More specifically, we compare \textit{DKMO} with \textit{kernel SVM}, and \textit{M-DKMO} with \textit{Uniform} and \textit{UFO-MKL} respectively to demonstrate that one can achieve better performance by replacing the conventional kernel learning strategies with the proposed deep optimization. We adopt flowers17 and flowers102 \footnote{\url{www.robots.ox.ac.uk/~vgg/data/flowers}}, two standard benchmarking datasets for image classification with kernel methods. Both datasets are comprised of flower images belonging to $17$ and $102$ categories respectively. The precomputed $\chi^2$ distance matrices were calculated based on bag of visual words of features such as HOG, HSV, SIFT etc. The variety of attributes enables the evaluation of different fusion algorithms: a large class of features that characterize colors, shapes and textures can be exploited while discriminating between different image categories \cite{jhuo2010boosted, bucak2014multiple}. 
%In object recognition and scene understanding , combining multiple feature representations has been shown to yield significantly higher classification accuracies. The reason behind this success is intuitive: 

We construct $\chi^2$ kernels from these distance matrices as $k(\x_i,\x_j)=e^{-\gamma l(\x_i,\x_j)}$, where $l$ denotes the distance between $\x_i$ and $\x_j$. Following \cite{orabona2012multi}, the $\gamma$ value is empirically estimated as the inverse of the average pairwise distances. To be consistent with the setting from \cite{qi2014pairwise} on the flowers102 dataset, we consider training on both $20$ samples per class and $30$ samples per class respectively. The experimental results for single kernels are shown in Figure \ref{fig:res:flowers} and results for multiple kernel fusion are shown in Table \ref{table:img}, where we measure the classification accuracy as the averaged fraction of correctly predicted labels among all classes. As can be seen, \textit{DKMO} achieves competitive or better accuracy on all single kernel cases and \textit{M-DKMO} consistently outperforms \textit{UFO-MKL}. In many cases the improvements are significant, for example kernel $6$ in the flowers17 dataset, kernel $1$ in the flowers102 dataset and the multiple kernel fusion result for the flowers17 dataset.

\subsection{Protein Subcellular Localization - Lack of Explicit Feature Sources}
\label{sec:exp:only_kernels}

%\begin{figure*}[!h]
%	\centering
%	\subfloat[Plant]{\includegraphics[width=0.3\linewidth]{protein_0.png}}
%	\label{fig:protein1}
%	\subfloat[Non-plant]{\includegraphics[width=0.3\linewidth]{protein_1.png}}
%	\label{fig:protein2}\\
%	\subfloat[Psort$+$]{\includegraphics[width=0.3\linewidth]{protein_2.png}}
%	\label{fig:protein3}
%	\subfloat[Psort$-$]{\includegraphics[width=0.3\linewidth]{protein_3.png}}
%	\label{fig:protein4}
%	\caption{Single Kernel Performance on Protein Subcellular Datasets}
%	\label{fig:res:protein}
%\end{figure*}
\begin{figure*}[!h]
	\centering
	\subfloat[Plant]{\includegraphics[width=0.245\linewidth]{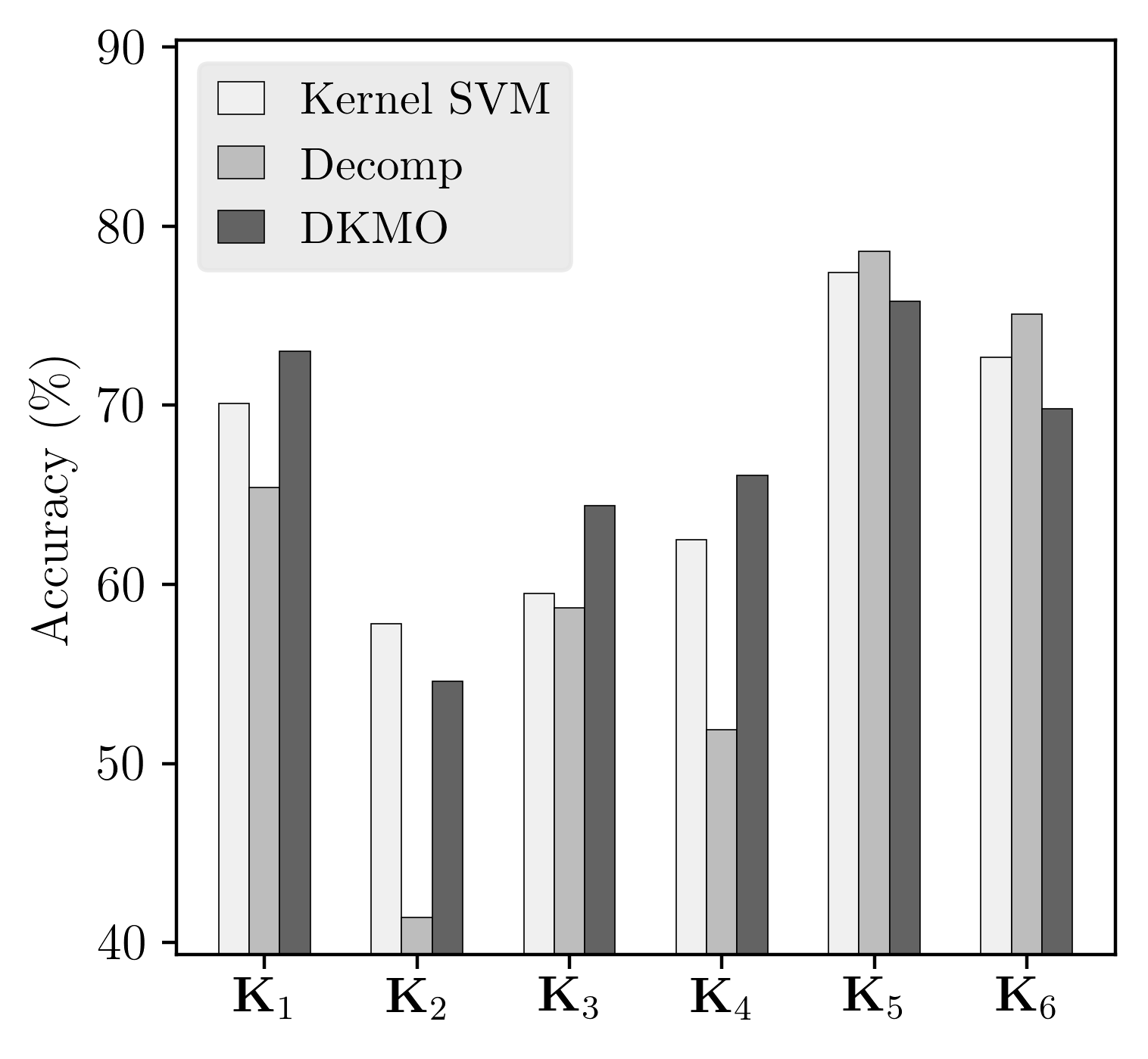}}
	\label{fig:protein1}
	\subfloat[Non-plant]{\includegraphics[width=0.245\linewidth]{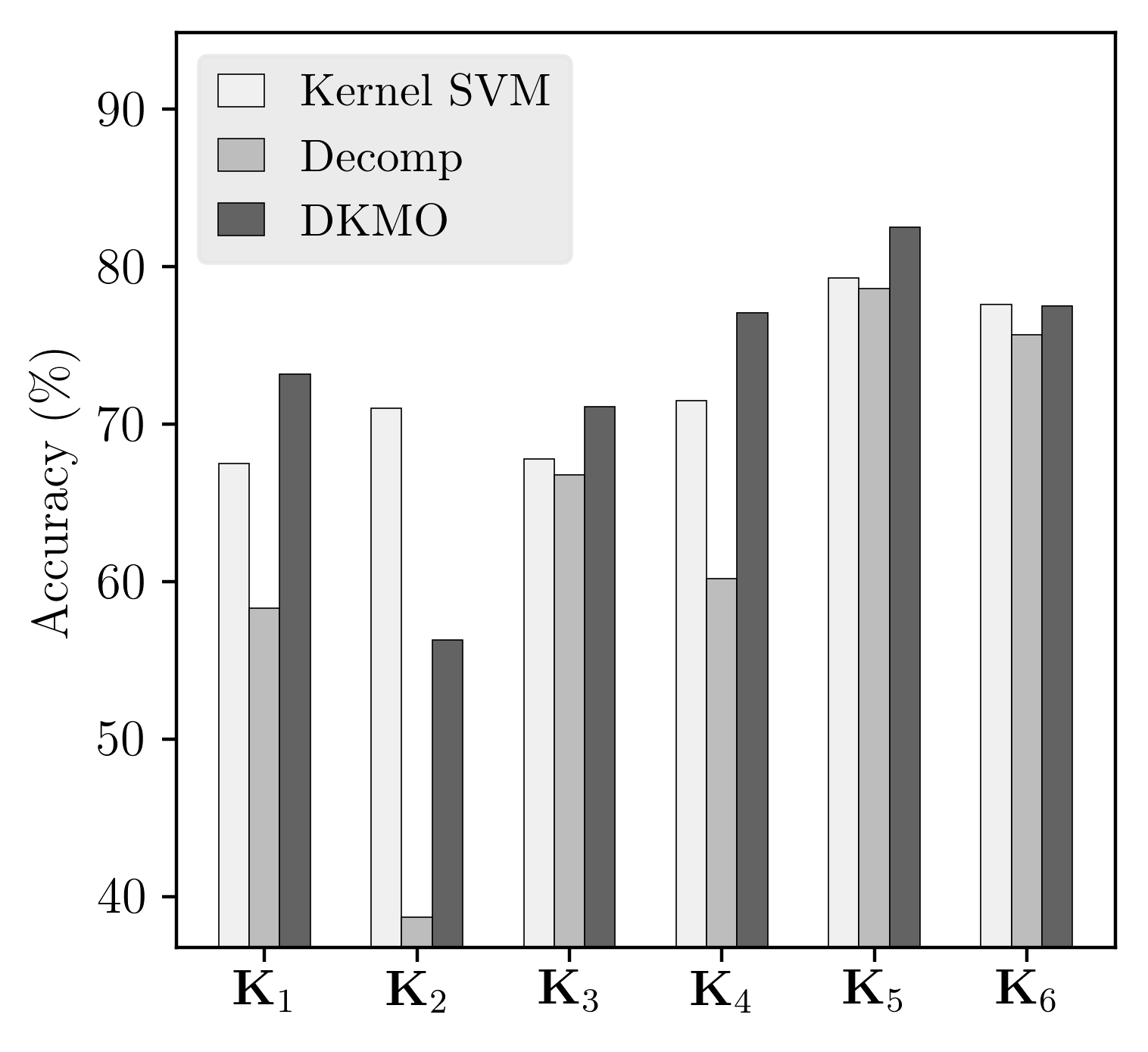}}
	\label{fig:protein2}
	\subfloat[Psort$+$]{\includegraphics[width=0.245\linewidth]{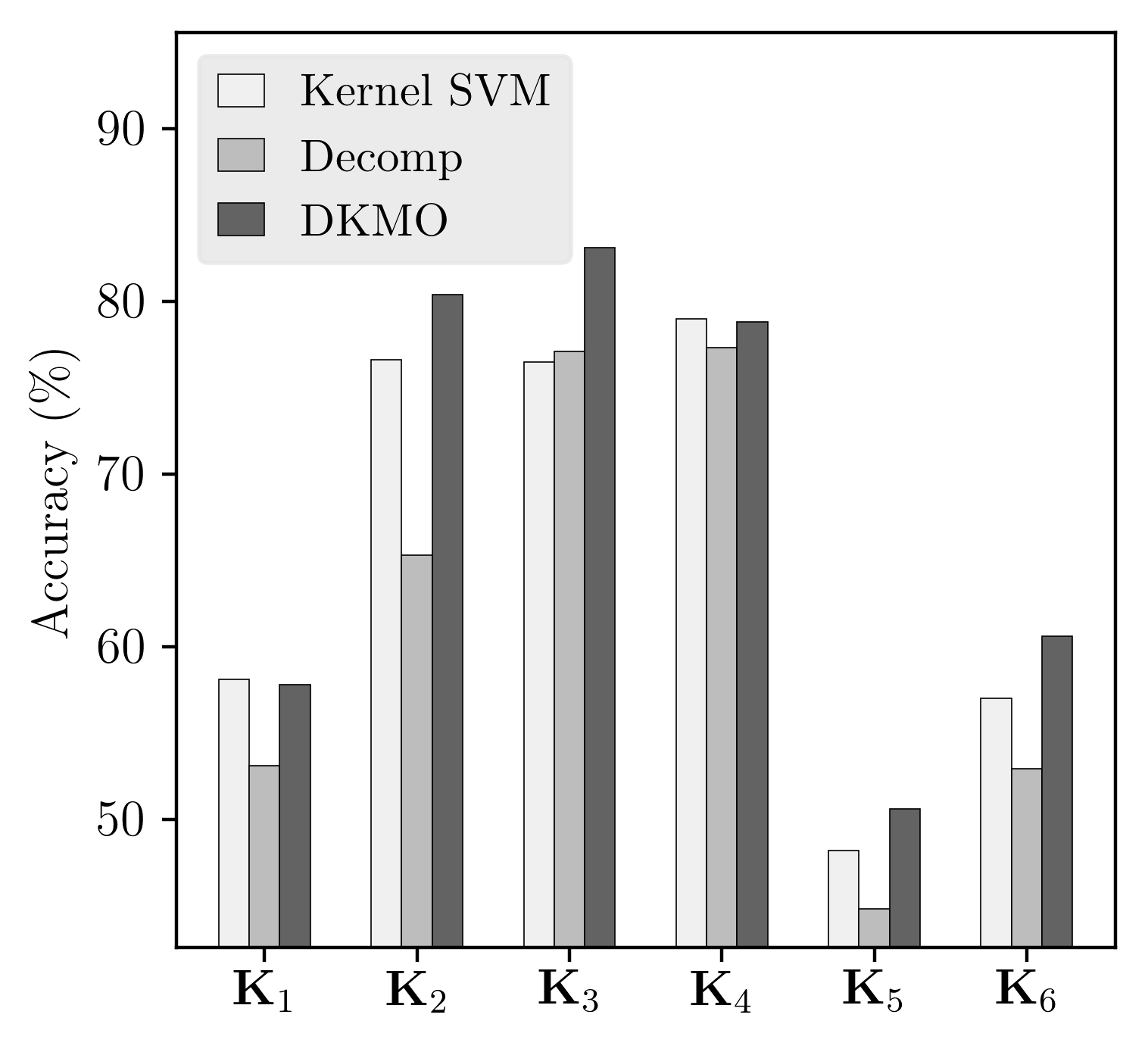}}
	\label{fig:protein3}
	\subfloat[Psort$-$]{\includegraphics[width=0.245\linewidth]{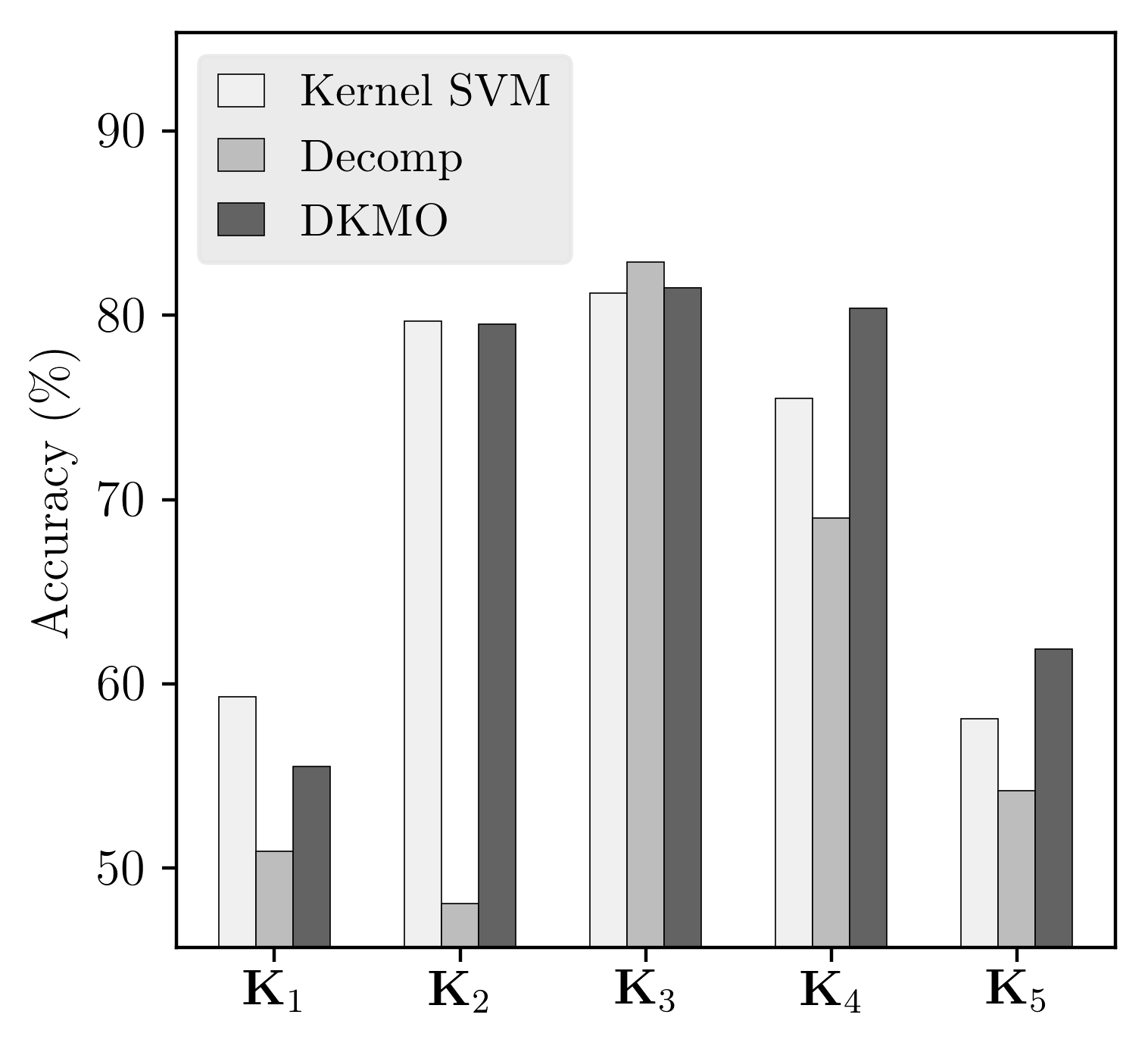}}
	\label{fig:protein4}
	\caption{Single Kernel Performance on Protein Subcellular Datasets}
	\label{fig:res:protein}
\end{figure*}

% Multiple kernels table for flowers datasets.
\begin{table}[t]
	\caption{Multiple Kernel Fusion Performance on Flowers Datasets}
	\label{table:img}
	\setlength\tabcolsep{2.5pt}
	\centering
	\begin{tabularx}{0.8\linewidth}{| Y Y Y |}
		\hline
		\textbf{Uniform} & \textbf{UFO-MKL} & \textbf{M-DKMO} \\
		\hline
		\hline
		\rowcolor{Gray}
		\multicolumn{3}{|c|}{\textsc{Flowers17}, $n=1360$} \\
		\hline
		85.3 & 87.1 & \textbf{90.6} \\
		\hline
		\rowcolor{Gray}
		\multicolumn{3}{|c|}{\textsc{Flowers102 - 20}, $n=8189$} \\
		\hline
		69.9 & 75.7 & \textbf{76.5} \\
		\hline
		\rowcolor{Gray}
		\multicolumn{3}{|c|}{\textsc{Flowers102 - 30}, $n=8189$} \\
		\hline
		73.0 & 80.4 & \textbf{80.7} \\
		\hline
	\end{tabularx}
\end{table}

\begin{figure}[t]
	\centering
	\subfloat[\textit{Decomp}]{\includegraphics[width=0.48\linewidth]{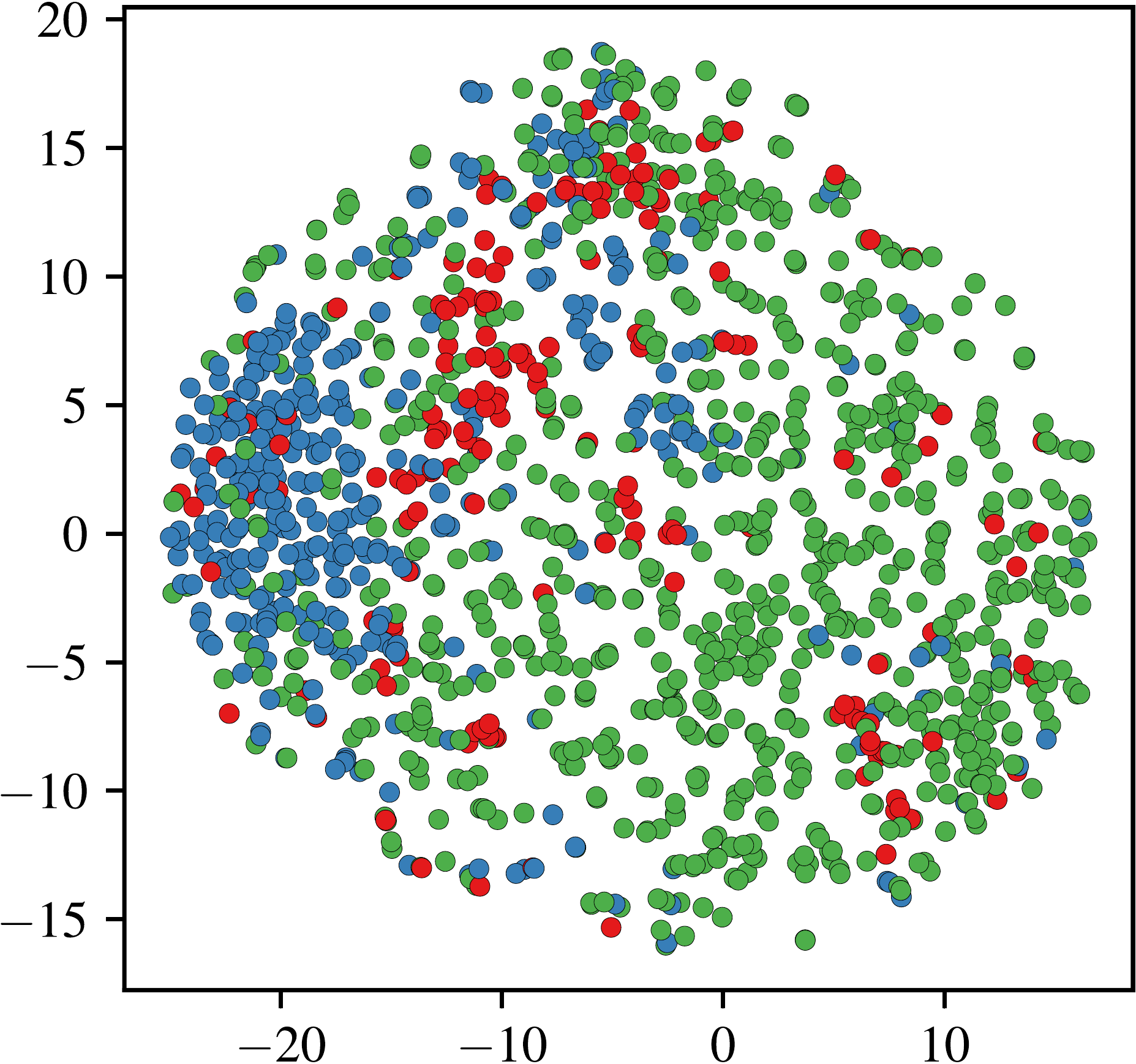}}
	\label{fig:tsne-a}\hfill
	\subfloat[Proposed \textit{DKMO}]{\includegraphics[width=0.48\linewidth]{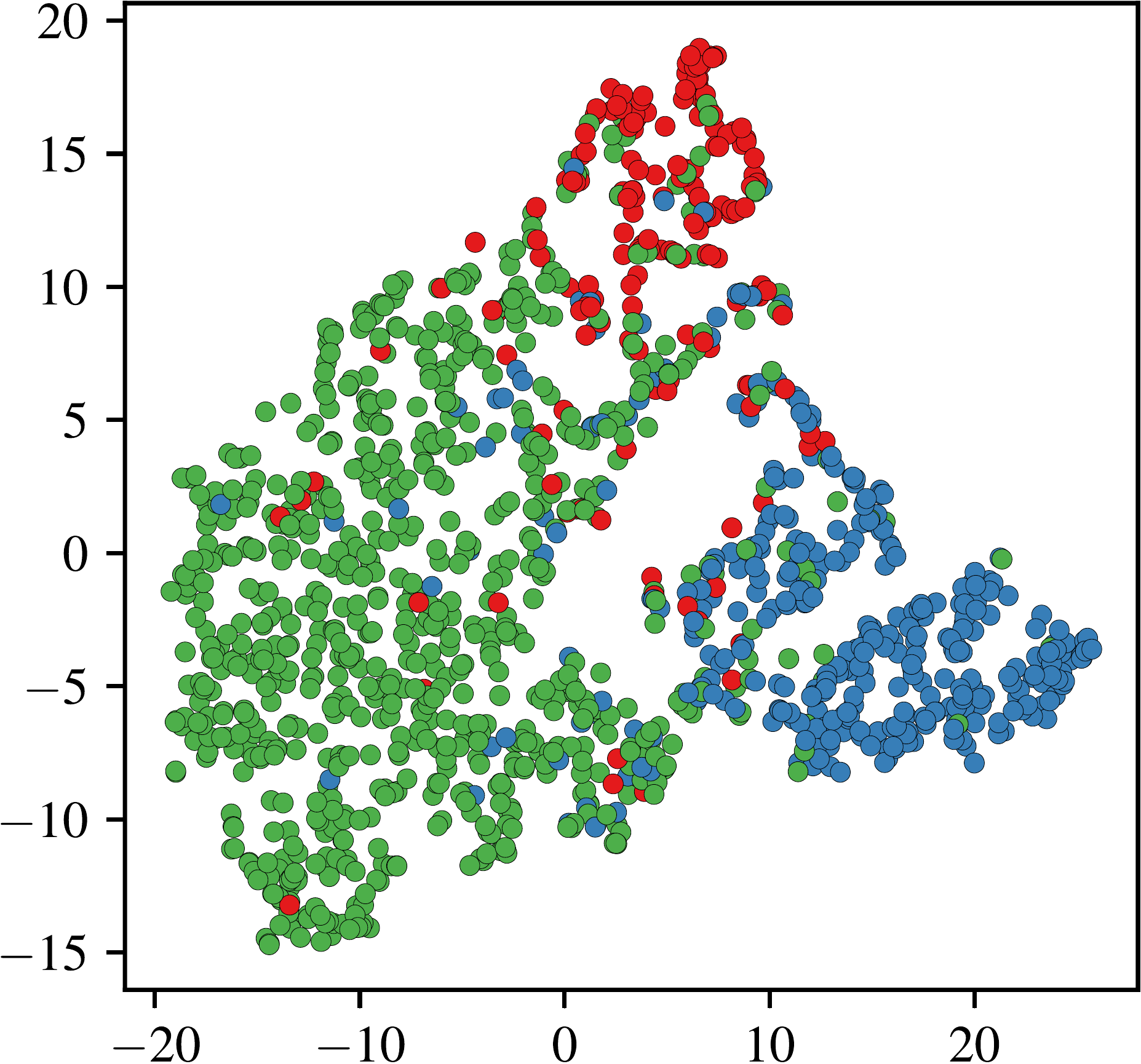}}
	\label{fig:tsne-b}\\
	\subfloat[\textit{Uniform}]{\includegraphics[width=0.48\linewidth]{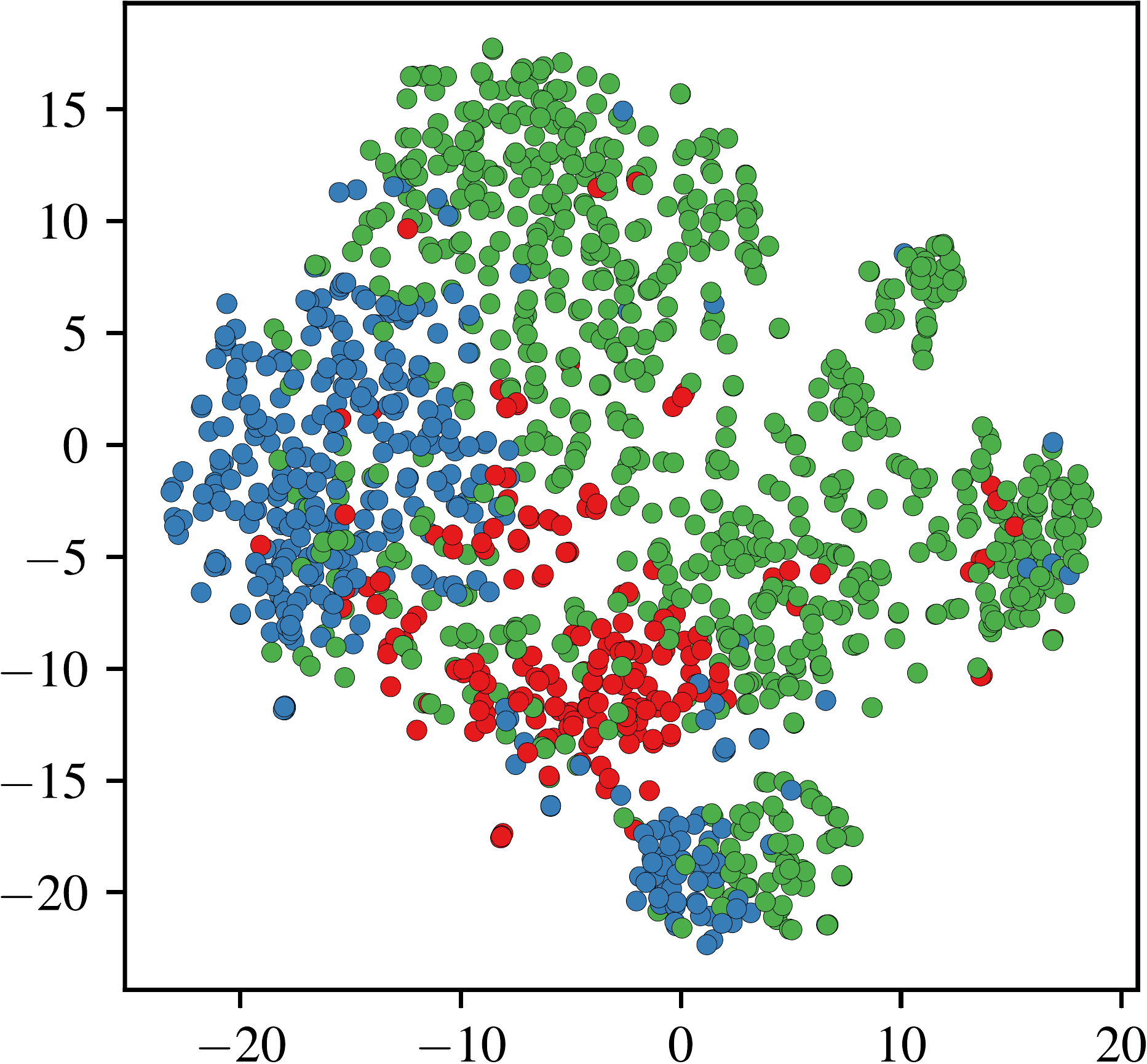}}
	\label{fig:tsne-c}\hfill
	\subfloat[Proposed \textit{M-DKMO}]{\includegraphics[width=0.48\linewidth]{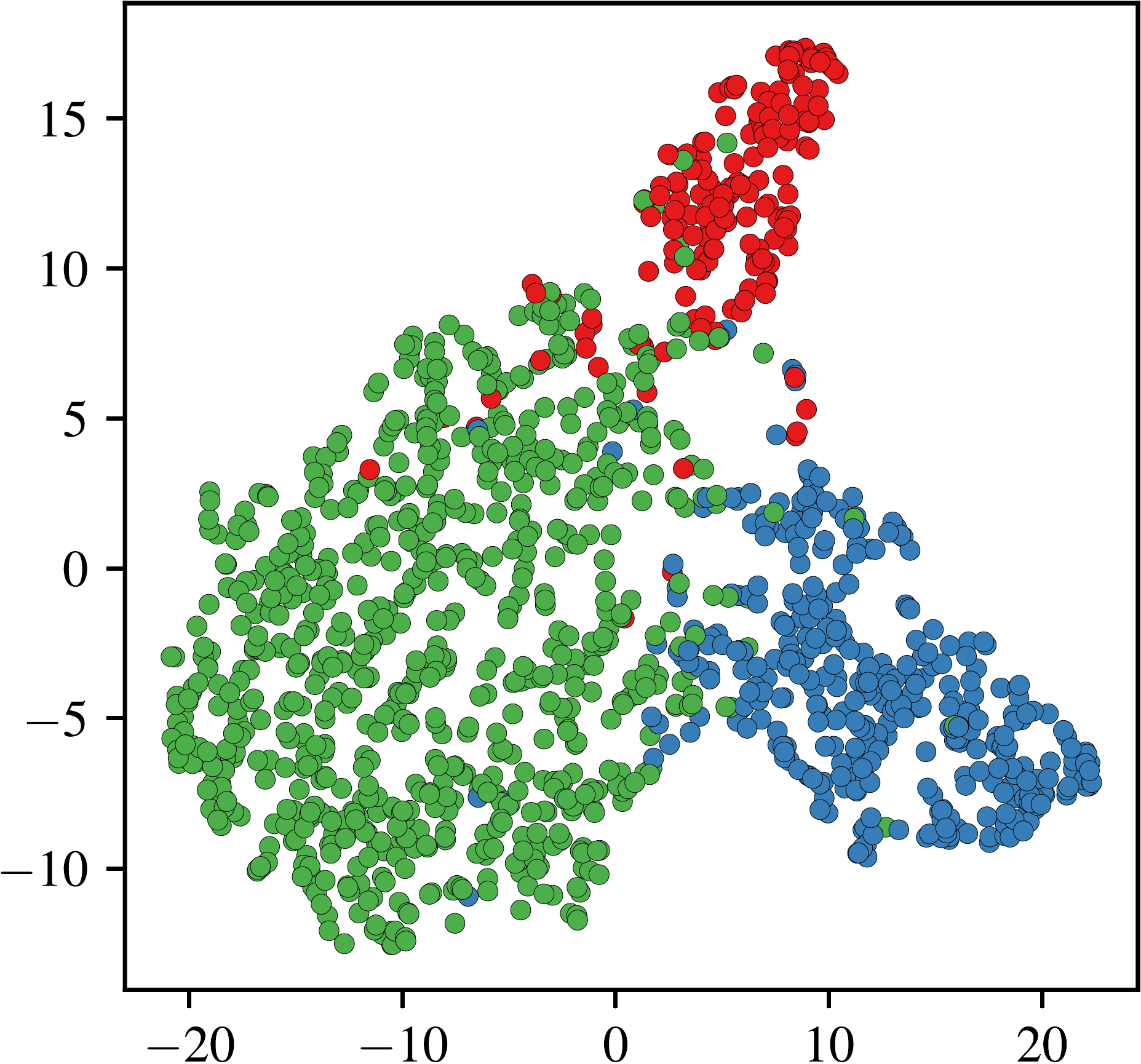}}
	\label{fig:tsne-d}
	\caption{$2$D T-SNE visualizations of the representations obtained for the non-plant dataset using the base kernel (Kernel $5$), uniform multiple kernel fusion, and the learned representations from \textit{DKMO} and \textit{M-DKMO}. The samples are colored by their corresponding class associations.}
	\label{fig:tsne}
\end{figure}

In this section, we consider the case where features are not directly available from data. This is a common scenario for many problems in bioinformatics, where conventional kernel methods have been successfully applied \cite{ong2008automated,ding2001multi,andreeva2014scop2}. More specifically, we focus on predicting the protein subcellular localization from protein sequences. We use $4$ datasets from \cite{ong2008automated} \footnote{\url{www.raetschlab.org/suppl/protsubloc}}: plant, non-plant, psort$+$ and psort$-$ belonging to $3-5$ classes. Among the $69$ sequence motif kernels, we sub-select $6$, which encompass all $5$ patterns for each substring format (except for psort$-$, where one invalid kernel is removed). Following standard practice, a $50-50$ random split is performed to obtain the train and test sets. Since explicit feature sources are not available, the dense embeddings are obtained using the conventional Nystr\"{o}m sampling method. 

The experimental results are shown in Figure \ref{fig:res:protein} and Table \ref{table:bio}. The first observation is that for \textit{Decomp}, although the optimal decomposition is used to obtain the features, the results are still inferior and inconsistent. This demonstrates that under such circumstances when features are not accessible, it is necessary to work directly from kernels and build the model. Second, we observe that on all datasets, \textit{DKMO} consistently produces improved or at least similar classification accuracies in comparison to the baseline \textit{kernel SVM}. For the few cases where \textit{DKMO} is inferior, for example kernel $2$ in non-plant, the quality of the Nystr\"{o}m approximation seemed to be the reason. By adopting more sophisticated approximations, or increasing the size of the ensemble, one can possibly make \textit{DKMO} more effective in such scenarios. Furthermore, in the multiple kernel learning case, the proposed \textit{M-DKMO} approach produces improved performance consistently.

Finally, in order to understand the behavior of the representations generated by different approaches, we employ the t-SNE algorithm \cite{maaten2008visualizing} and obtain $2$-D visualizations of the considered baselines and the proposed approaches (Figure \ref{fig:tsne}). For demonstration, we consider the representation from \textit{Decomp} of kernel $5$ and \textit{Decomp} of the kernel from \textit{Uniform} in the non-plant dataset. In both \textit{DKMO} and \textit{M-DKMO}, we performed t-SNE on the representation obtained from the fusion layers. The comparisons in the Figure \ref{fig:tsne} show that the proposed single kernel learning and kernel fusion methods produce highly discriminative representations than the corresponding conventional approaches.

\subsection{Sensor-based Activity Recognition - Limited Data Case}
\label{sec:exp:limited}

\begin{figure*}[t]
	\centering
	\centerline{\includegraphics[width=0.9\textwidth]{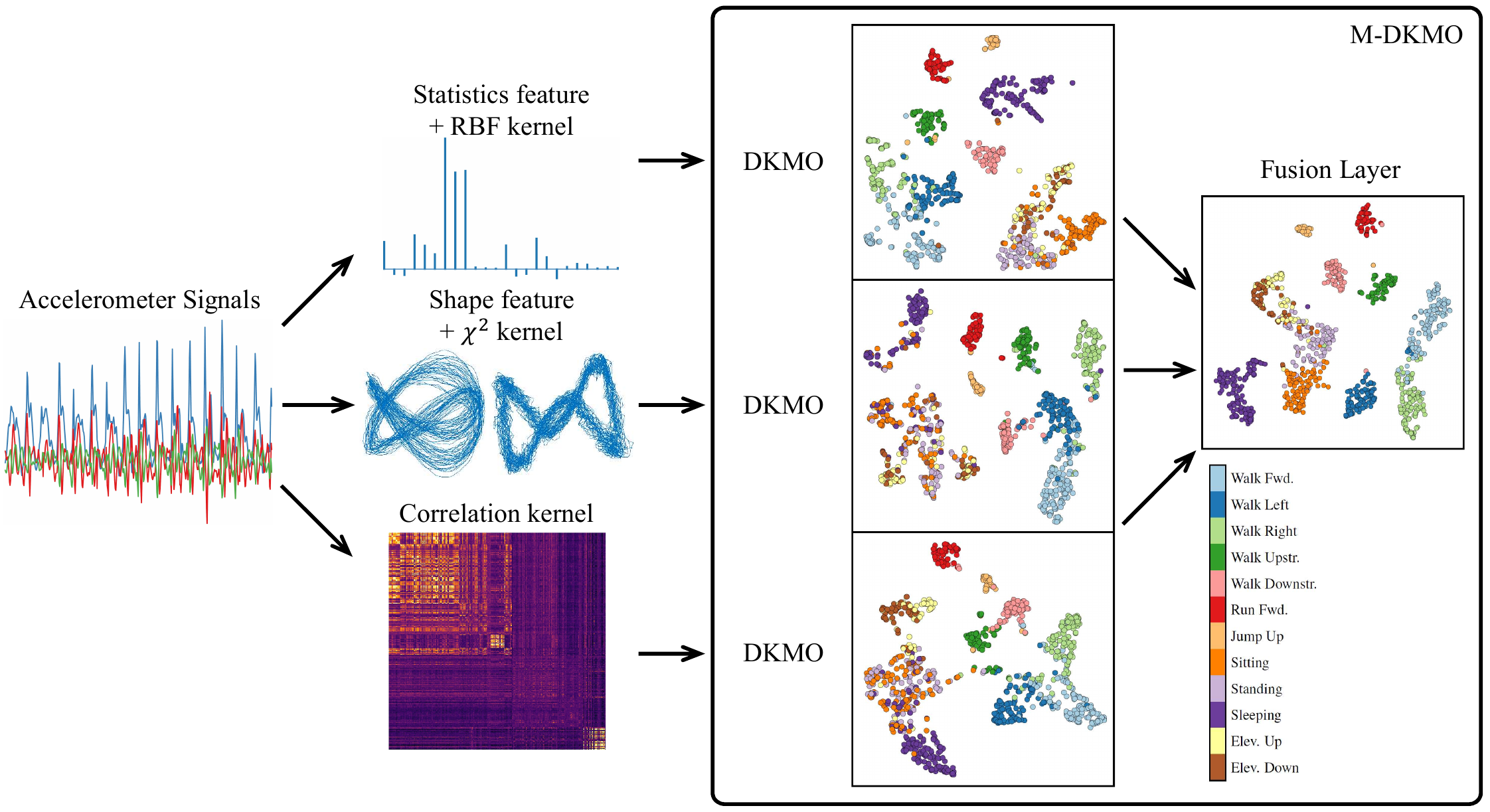}}
	\caption{Visualization of proposed framework applied on USD-HAD dataset: we show the raw $3$-axis accelerometer signal and extracted $3$ distinct types of features: the time-series statistics, topological structure where we extract TDE descriptors and the correlation kernel. Furthermore, we show the t-SNE visualization of the representations learned by \textit{DKMO} and \textit{M-DKMO}, where all points are classes coded according to the colorbar.}
	\label{fig:act_framework}
\end{figure*} 

% Table with only multiple kernel results.
\begin{table}[t]
	\caption{Multiple Kernel Fusion Performance on Protein Subcellular Datasets}
	\setlength\tabcolsep{5.5pt}
	\centering
	\begin{tabularx}{0.8\linewidth}{| Y Y Y Y |}
		\hline
		\textbf{Concat} & \textbf{Uniform} & \textbf{UFO-MKL} & \textbf{M-DKMO} \\
		\hline
		\hline
		\rowcolor{Gray}
		\multicolumn{4}{|c|}{\textsc{Plant}, $n=940$} \\
		\hline
		90.4 & 90.3 & 90.4 & \textbf{90.9} \\
		\hline
		\rowcolor{Gray}
		\multicolumn{4}{|c|}{\textsc{Non-plant}, $n=2732$} \\
		\hline
		88.4 & 91.1 & 90.3 & \textbf{93.8} \\
		\hline
		\rowcolor{Gray}
		\multicolumn{4}{|c|}{\textsc{Psort$+$}, $n=541$} \\
		\hline
		80.6 & 80.1 & \textbf{82.8} & 82.4 \\
		\hline
		\rowcolor{Gray}
		\multicolumn{4}{|c|}{\textsc{Psort$-$}, $n=1444$} \\
		\hline
		82.5 & 85.7 & \textbf{89.1} & 87.2 \\
		\hline
	\end{tabularx}
	\label{table:bio}
\end{table}

\begin{figure}[!h]
	\centering
	\centerline{\includegraphics[width=0.65\linewidth]{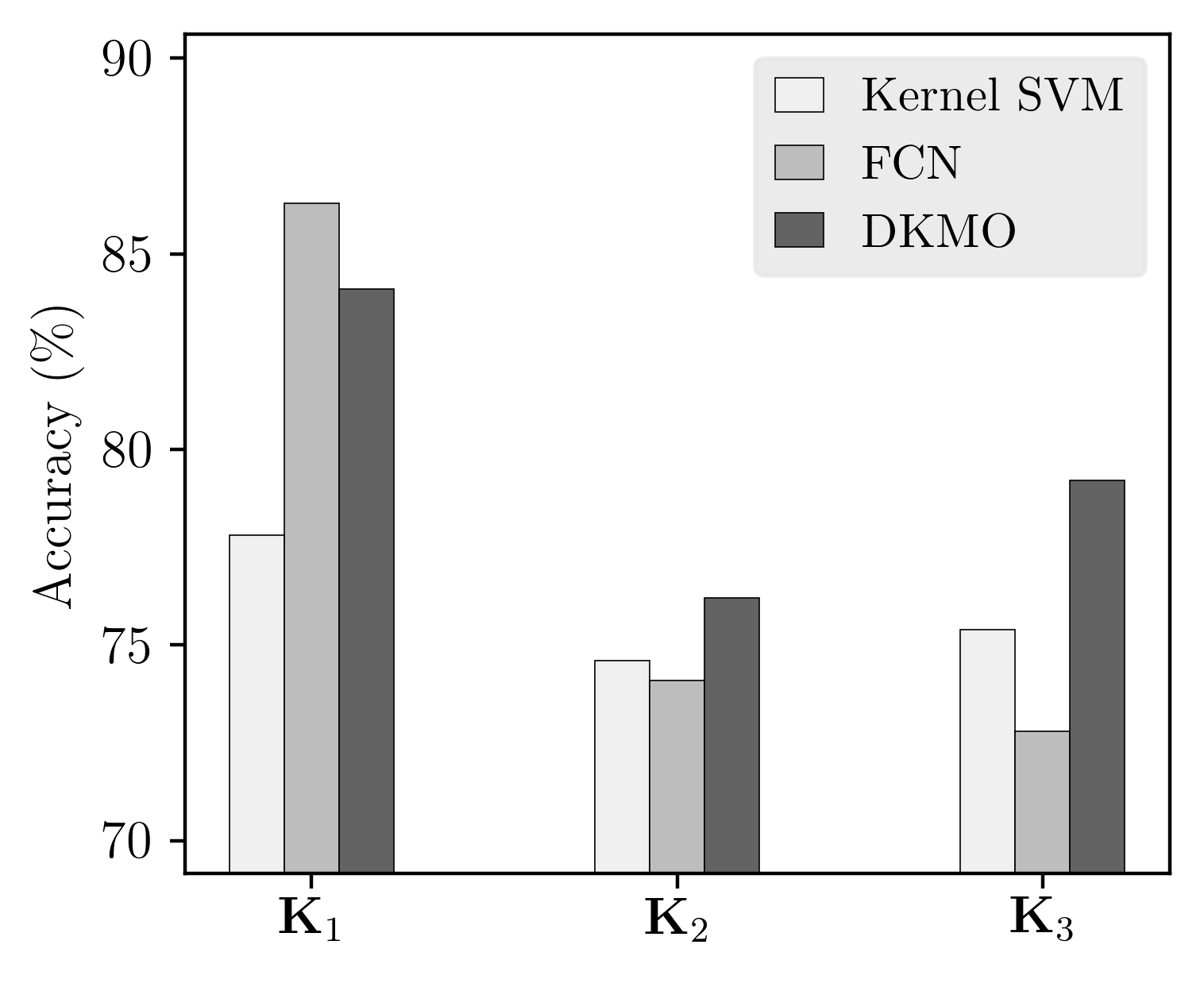}}
	\caption{Single Kernel Performance on USC-HAD Datasets}
	\label{fig:res:activity}
\end{figure}

% Original table for activity dataset
%\begin{table*}[h]
%	\setlength\tabcolsep{2.5pt}
%	\caption{Classification Performance on USC-HAD Datasets}
%	\centering
%	\begin{tabularx}{0.7\textwidth}{| c Y Y Y | c c c c|}
%		\hline
%		\multicolumn{4}{|c|}{\textbf{Single Kernel Learning}} & \multicolumn{4}{c|}{\textbf{Multiple Kernel Learning}} \\
%		Method & Kernel 1 & Kernel 2 & Kernel 3 & Uniform & UFO-MKL & FCN & M-DKMO \\
%		\hline
%		\hline
%		\rowcolor{Gray}
%		\multicolumn{8}{|c|}{\textsc{USC-HAD}, $n=5353$} \\
%		\hline
%		Kernel SVM & 77.8 & 74.6 & 75.4 & \multirow{3}{*}{89.0} & \multirow{3}{*}{87.1} & \multirow{3}{*}{85.9} & \multirow{3}{*}{\textbf{90.4}}\\
%		FCN & 86.3 & 74.1 & 72.8 & & & & \\
%		DKMO & 84.1 & \textbf{76.2} & \textbf{79.2} & & & & \\
%		\hline
%	\end{tabularx}
%	\label{table:activity}
%\end{table*}

In this section, we focus on evaluating the performance of proposed architectures where training data are limited. A typical example under this scenario is sensor-based activity recognition, where the sensor time-series data have to be obtained from human subjects through long-term physical activities. For evaluation, we compare \textit{(M)-DKMO} with both \textit{FCN} and kernel learning algorithms.

Recent advances in activity recognition have demonstrated promising results in fitness monitoring and assisted living \cite{zhang2012usc}. However, when applied to smartphone sensors and wearables, existing algorithms still have limitations dealing with the measurement inaccuracies and noise. In \cite{song2016consensus}, the authors proposed to address this challenge by performing sensor fusion, wherein each sensor is characterized by multiple feature sources, which naturally enables multiple kernel learning schemes. 

We evaluate the performance of our framework using the USC-HAD dataset \footnote{\url{sipi.usc.edu/HAD}}, which contains $12$ different daily activities performed by each of the subjects. The measurements are obtained using a $3$-axis accelerometer at a sampling rate of $100$Hz. Following the standard experiment methodology, we extract non-overlapping frames of $5$ seconds each,  creating a total of $5353$ frames. We perform a $80-20$ random split on the data to generate the train and test sets. In order to characterize distinct aspects of the time-series signals, we consider $3$ sets of features:

\begin{enumerate}[1)]
\item Statistics feature including mean, median, standard deviation, kurtosis, skewness, total acceleration, mean-crossing rate and dominant frequency. These features encode the statistical characteristics of the signals in both time and frequency domains.

\item Shape feature derived from Time Delay Embeddings (TDE) to model the underlying dynamical system \cite{frank2010activity}. The TDEs of a time-series signal $\mathbf{x}$ can be defined as a matrix $\mathbf{S}$ whose $i$th row is $\mathbf{s}_i=[x_i,x_{i+\tau},\dots,x_{t+(d'-1)\tau}]$, where $d'$ is number of samples and $\tau$ is the delay parameter. The time-delayed observation samples can be considered as points in $\mathbb{R}^{d'}$, which is referred as the delay embedding space. In this experiment, the delay parameter $\tau$ is fixed to $10$ and embedding dimension $d'$ is chosen to be $8$. Following the approach in \cite{frank2010activity}, we use Principle Component Analysis (PCA) to project the embedding to $3$-D for noise reduction. To model the topology of the delayed observations in $3$-D, we measure the pair-wise distances between samples as $\|\mathbf{s}_i-\mathbf{s}_j\|_2$ \cite{venkataraman2016shape} and build the distance histogram feature with a pre-specified bin size.

\item Correlation features characterizing the dependence between time-series signals. We calculate the absolute value of the Pearson correlation coefficient. To account for shift between the two signals, the maximum absolute coefficient for a small range of shift values is identified. We ensure that the correlation matrix is a valid kernel by removing the negative eigenvalues. Given the eigen-decomposition of the correlation matrix $\mathbf{R} = \mathbf{U_R} \mathbf{\Lambda_R} \mathbf{U}_\mathbf{R}^T$, where $\mathbf{\Lambda_R}=\text{diag}(\sigma_1,\dots,\sigma_n)$ and $\sigma_1\geq\dots \geq\sigma_r\geq0\geq\sigma_{r+1}\geq...\geq\sigma_n$, the correlation kernel is constructed as $\K=\mathbf{U_R} \mathbf{\hat{\Lambda}_R} \mathbf{U}_\mathbf{R}^T$, where $\mathbf{\hat{\Lambda}_R}=\text{diag}(\sigma_1,\dots,\sigma_r,0,...,0)$.
\end{enumerate}

% Multiple kernel table for activity dataset.
\begin{table}[t]
	\setlength\tabcolsep{2.5pt}
	\caption{Multiple Kernel Fusion Performance on USC-HAD Datasets}
	\centering
	\begin{tabularx}{0.8\linewidth}{| Y Y Y Y |}
		\hline
		\textbf{Uniform} & \textbf{UFO-MKL} & \textbf{FCN} & \textbf{M-DKMO} \\
		\hline
		\hline
		\rowcolor{Gray}
		\multicolumn{4}{|c|}{\textsc{USC-HAD}, $n=5353$} \\
		\hline
		89.0 & 87.1 & 85.9 & \textbf{90.4}\\
		\hline
	\end{tabularx}
	\label{table:activity}
\end{table}

Figure \ref{fig:act_framework} illustrates the overall pipeline of this experiment. As it can be observed, the statistics and shape representations are explicit feature sources and hence the dense embeddings can be constructed using the clustered Nystr\"{o}m method (through RBF and $\chi^2$ kernel formulations respectively). On the other hand, the correlation representation is obtained directly based on the similarity metric and hence we employ the conventional Nystr\"{o}m approximations on the kernel. However, regardless of the difference in dense embedding construction, the kernel learning procedure is the same for both cases. From the t-SNE visualizations in Figure \ref{fig:act_framework}, we notice that the classes \textit{Sitting}, \textit{Standing}, \textit{Elevator Up} and \textit{Elevator Down} are difficult to discriminate using any of the individual kernels. In comparison, the fused representation obtained using the \textit{M-DKMO} algorithm results in a much improved class separation, thereby demonstrating the effectiveness of the proposed kernel fusion architecture. 

From the classification results in Figure \ref{fig:res:activity}, we observe that although \textit{FCN} obtains better result on the set of statistics features, it has inferior performance on shape and correlation features. On the contrary, \textit{DKMO} improves on \textit{kernel SVM} significantly for each individual feature set and is more consistent than \textit{FCN}. In the case of multiple kernel fusion in Table \ref{table:activity}, we have striking observations: 1) For \textit{FCN}, the fusion performance is in fact dragged down by the poor performance on shape and correlation features as in Figure \ref{fig:res:activity}. 2) The uniform merging of kernels is a very strong baseline and the state-of-the-art \textit{UFO-MKL} achieves lesser performance. 3) The proposed \textit{M-DKMO} framework further improves over uniform merging, thus evidencing its effectiveness in optimizing with multiple feature sources.

\section{Conclusions}
In this paper, we presented a novel approach to perform kernel learning using deep architectures. The proposed approach utilizes the similarity kernel matrix to generate an ensemble of dense embeddings for the data samples and employs end-to-end deep learning to infer task-specific representations. Intuitively, we learn representations describing the characteristics of different linear subspaces in the RKHS. By enabling the neural network to exploit the native space of a pre-defined kernel, we obtain models with much improved generalization. Furthermore, the kernel dropout process allows the predictive model to exploit the complementary nature of the different subspaces and emulate the behavior of kernel fusion using a backpropagation based optimization setting. In addition to improving upon the strategies adopted in kernel machine optimization, our approach demonstrates improvements over conventional kernel methods in different applications. We also showed that using these improved representations, one can also perform multiple kernel learning efficiently. In addition to showing good convergence characteristics, the M-DKMO approach consistently outperforms state-of-the-art MKL methods. The empirical results clearly evidence the usefulness of using deep networks as an alternative approach to building kernel machines. From another viewpoint, similar to the recent approaches such as the convolutional kernel networks \cite{mairal2016end}, principles from kernel learning theory can enable the design of novel training strategies for neural networks. This can be particularly effective in applications that employ fully connected networks and in scenarios where training data is limited, wherein bridging these two paradigms can lead to capacity-controlled modeling for better generalization.

%The critical part is dense embedding, which connects kernel methods to deep learning. To obtain effective dense embedding, we propose to use multiple Nystr\"{o}m approximate mappings and further use deep architecture to transform and fuse them. Moreover, we showed that with the improved kernel representations, the problem of multiple kernel fusion can be better solved under the similar deep learning setting. As we have demonstrated in experiments, the proposed framework is general in that it can be applied to a wide range of problems where kernels are properly defined. The framework is also flexible so that more advanced techniques and new ideas can be easily incorporated into different parts. For example, we plan to investigate on obtaining better Nystr\"{o}m approximate mappings and merge the step into the deep architecture. Besides, research can also be performed on new neural network architectures which align with kernel structures. 

\bibliographystyle{IEEEtran}
\bibliography{IEEEabrv,refs}

%\begin{IEEEbiography}{Huan~Song}
%\end{IEEEbiography}
%
%\begin{IEEEbiography}{Jayaraman~J.~Thiagarajan}
%\end{IEEEbiography}
%
%\begin{IEEEbiography}{Prasanna~Sattigeri}
%\end{IEEEbiography}
%
%\begin{IEEEbiography}{Andreas~Spanias}

\end{document}